\documentclass[lettersize,journal]{IEEEtran}
\usepackage{amsmath,amsfonts}
\usepackage{graphicx}
\usepackage{amssymb}
\usepackage{array}
\usepackage[caption=false,font=normalsize,labelfont=sf,textfont=sf]{subfig}
\usepackage{textcomp}
\usepackage{threeparttable}
\usepackage{makecell}
\usepackage{color}
\usepackage{stfloats}
\usepackage{wasysym}
\usepackage{url}
\usepackage{verbatim}
\usepackage{graphicx}
\usepackage{cite}
\usepackage{multirow}
\usepackage{hhline}
\usepackage{booktabs}
\usepackage{hyperref}
\hypersetup{
	breaklinks=true,
	letterpaper=blue,
	bookmarks=false     
}
\usepackage{pifont}
\usepackage[ruled,linesnumbered]{algorithm2e}
\SetKwInput{KwInput}{Input}
\SetKwInput{KwOutput}{Output} 
\begin{document}

\title{SamLP: A Customized Segment Anything Model for License Plate Detection}

\author{Haoxuan~Ding,
	Junyu~Gao,~\IEEEmembership{Member,~IEEE,}
	Yuan~Yuan,~\IEEEmembership{Senior~Member,~IEEE,}
	and~Qi~Wang,~\IEEEmembership{Senior~Member,~IEEE}
	\thanks{H. Ding is with the Unmanned System Research Institute, and with the School of Artificial Intelligence, Optics and Electronics (iOPEN), Northwestern Polytechnical University, Xi’an 710072, P. R. China. (e-mail: haoxuan.ding@mail.nwpu.edu.cn)}%
	\thanks{J. Gao, Y. Yuan, and Q. Wang are with the School of Artificial Intelligence, Optics and Electronics (iOPEN), Northwestern Polytechnical University, Xi'an 710072, P.R. China. (e-mail: gjy3035@gmail.com, y.yuan1.ieee@gmail.com, crabwq@gmail.com)}%
	\thanks{Q. Wang is the corresponding author.}
}


\markboth{IEEE TRANSACTIONS ON INTELLIGENT TRANSPORTATION SYSTEMS}%
{Ding \MakeLowercase{\textit{et al.}}: SamLP: A Customized Segment Anything Model for License Plate Detection}

\IEEEpubid{0000--0000/00\$00.00~\copyright~2021 IEEE}

\maketitle

\begin{abstract}
With the emergence of foundation model, this novel paradigm of deep learning has encouraged many powerful achievements in natural language processing and computer vision. There are many advantages of foundation model, such as excellent feature extraction power, mighty generalization ability, great few-shot and zero-shot learning capacity, \emph{etc.} which are beneficial to vision tasks. As the unique identity of vehicle, different countries and regions have diverse license plate (LP) styles and appearances, and even different types of vehicles have different LPs. However, recent deep learning based license plate detectors are mainly trained on specific datasets, and these limited datasets constrain the effectiveness and robustness of LP detectors. To alleviate the negative impact of limited data, an attempt to exploit the advantages of foundation model is implement in this paper. We customize a vision foundation model, \emph{i.e.} Segment Anything Model (SAM), for LP detection task and propose the first LP detector based on vision foundation model, named SamLP. Specifically, we design a Low-Rank Adaptation (LoRA) fine-tuning strategy to inject extra parameters into SAM and transfer SAM into LP detection task. And then, we further propose a promptable fine-tuning step to provide SamLP with prompatable segmentation capacity. The experiments show that our proposed SamLP achieves promising detection performance compared to other LP detectors. Meanwhile, the proposed SamLP has great few-shot and zero-shot learning ability, which shows the potential of transferring vision foundation model. The code is available at \url{https://github.com/Dinghaoxuan/SamLP}.
\end{abstract}
\begin{IEEEkeywords}
	Automatic license plate detection, Vision foundation model, Segment Anything Model, Parameter-efficient fine-tuning
\end{IEEEkeywords}

\section{Introduction}\label{sec:intro}

\begin{figure}[!h]
	\centering
	\subfloat{
		\label{fig0_0}
		\includegraphics[width=0.49\linewidth]{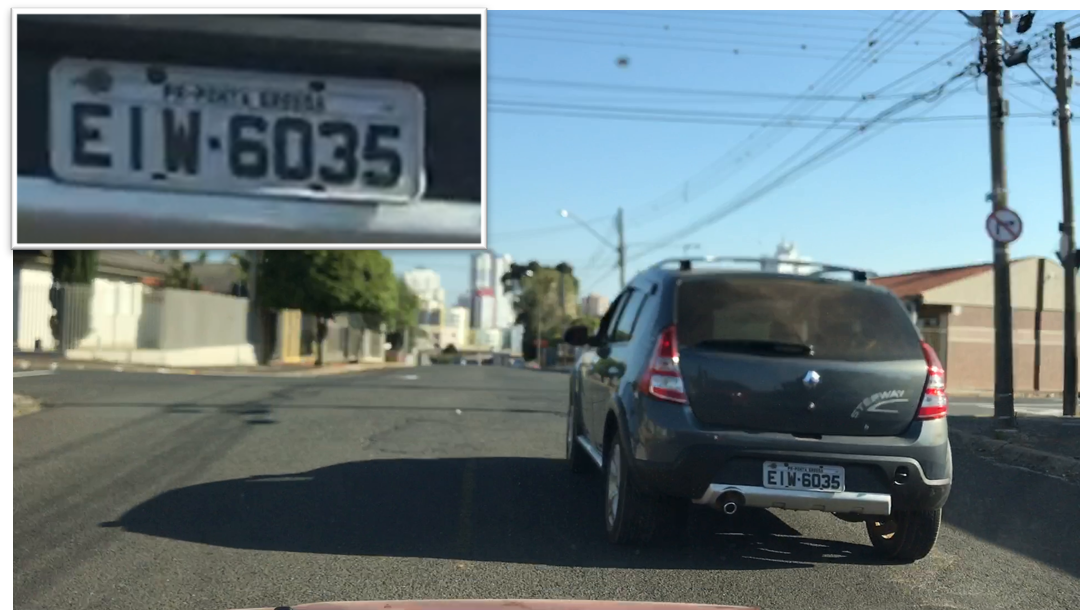}
		\includegraphics[width=0.49\linewidth]{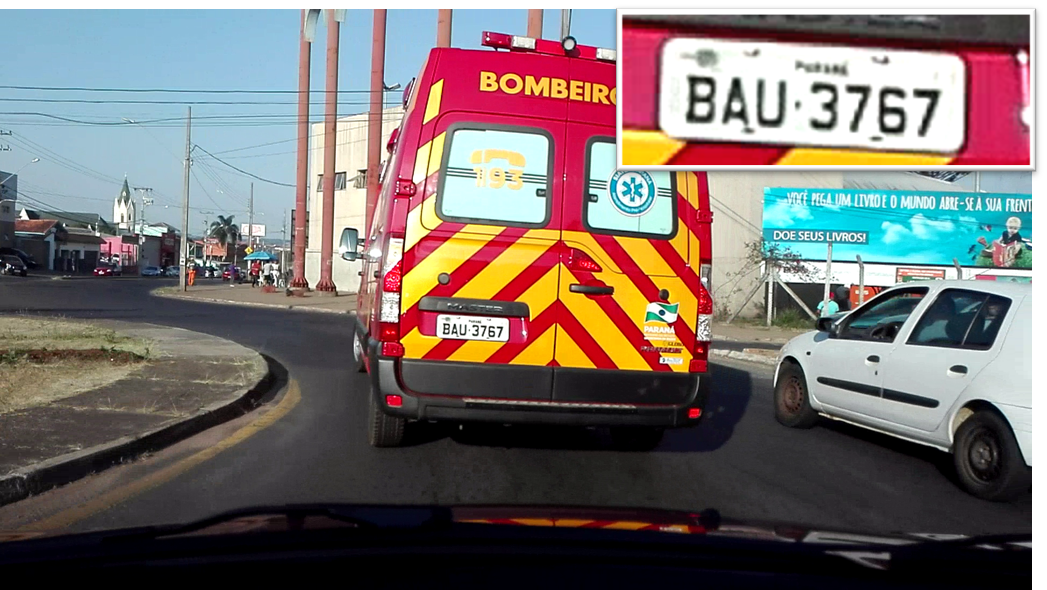}
	}
	\hfil
	\subfloat{
		\label{fig0_2}
		\includegraphics[width=0.49\linewidth]{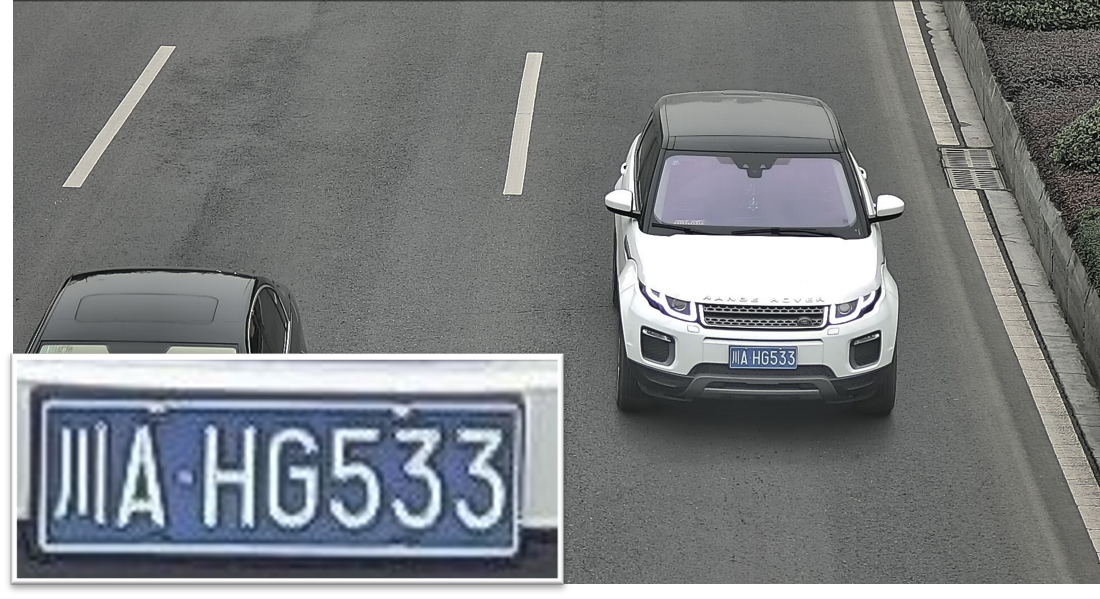}
		\includegraphics[width=0.49\linewidth]{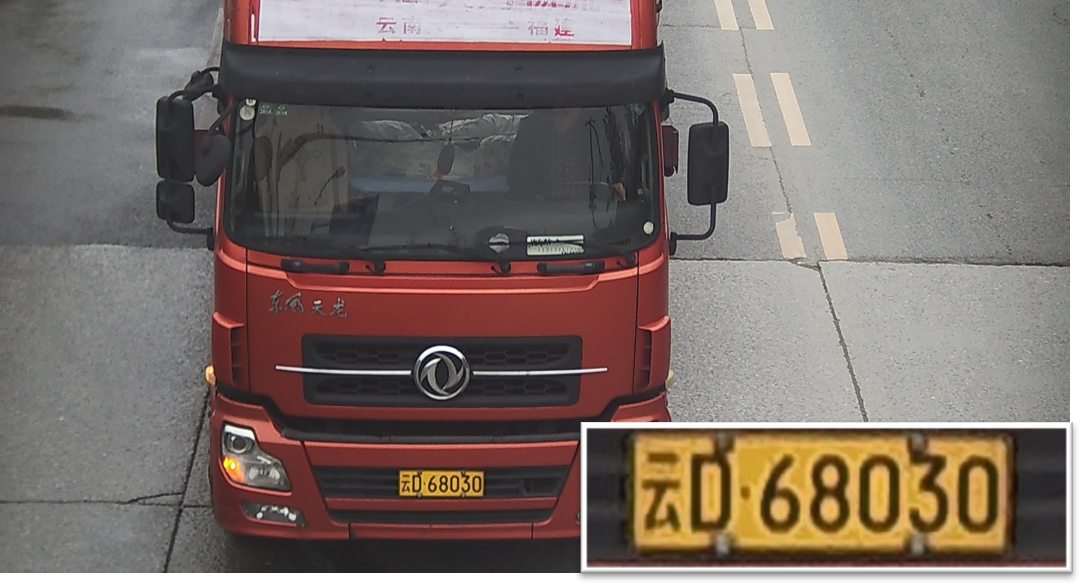}
	}
	\hfil
	\subfloat{
		\label{fig0_1}
		\includegraphics[width=0.49\linewidth]{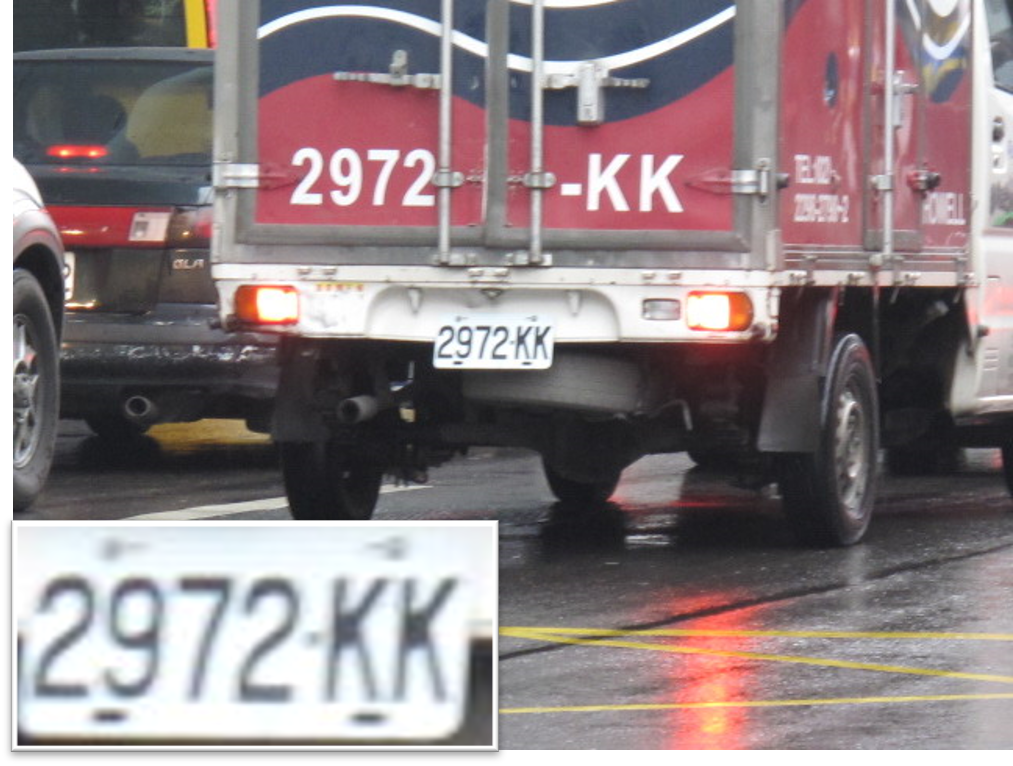}
		\includegraphics[width=0.49\linewidth]{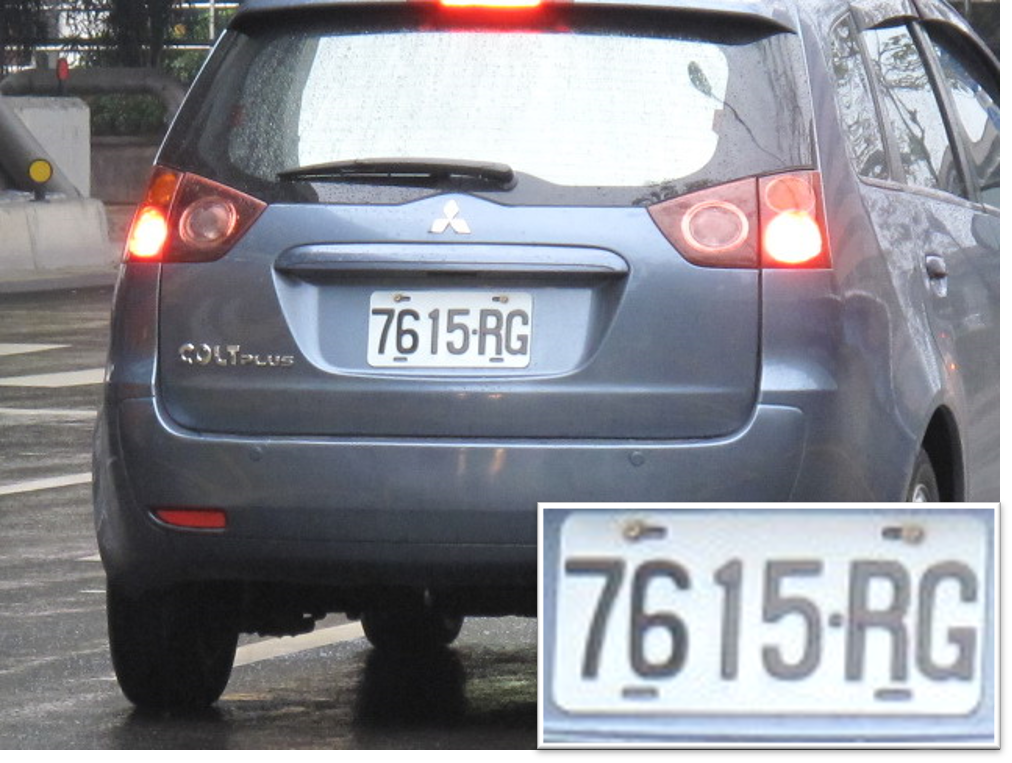}
	}
	\caption{The diverse license plates in the world. Different countries and regions have different license plate appearances and styles.}
	\label{fig_LP}
\end{figure}

\IEEEPARstart{F}{oundation} models \cite{DBLP:journals/corr/abs-2108-07258}, an emerging paradigm in deep learning, has received widespread attention in artificial intelligence. Foundation models has achieved promising performance in nature language processing (NLP) and vision language (VL), such as ChatGPT \cite{ChatGPT}, GPT-4 \cite{GPT4}, CLIP \cite{CLIP}, \emph{etc}. In computer vision (CV), Segment Anything Model (SAM) \cite{SAM} is the first vision foundation model and it predicts the potential masks at given positions guided by points, bounding boxes, and masks (the details of SAM \cite{SAM} will be described in Section \ref{sec:method:SAM}). SAM \cite{SAM} has been introduced into many computer vision tasks, such as medical image segmentation \cite{DBLP:journals/corr/abs-2304-12306, DBLP:journals/corr/abs-2304-13785, DBLP:journals/corr/abs-2304-12620}, video object tracking and segmentation \cite{DBLP:journals/corr/abs-2304-11968, DBLP:journals/corr/abs-2305-12659}, 2D$/$3D object detection \cite{DBLP:journals/corr/abs-2304-09148, DBLP:journals/corr/abs-2306-02245}, image editing \cite{DBLP:journals/corr/abs-2304-06790, DBLP:journals/corr/abs-2305-15094}. The widespread application of SAM \cite{SAM} in above tasks mainly benefits from the excellent transfer and generalization ability of foundation model. These advantages of SAM \cite{SAM} will facilitate the vision tasks in intelligent transportation systems (ITS) effectively. In this paper, we attempt to utilize the characteristics of SAM \cite{SAM} to alleviate some problems in License Plate (LP) detection task, boosting the performance of LP detection. 
\IEEEpubidadjcol
As a unique identity of vehicle, the types of license plates are diverse. As shown in Fig. \ref{fig_LP}, different types of vehicles have different LPs, and the LPs from different countries or regions are also different. Meanwhile, different scenes apply different devices (\emph{e.g.} dashboard camera, video surveillance, \emph{etc}) to capture the images of LPs. Laroca \emph{et al.} \cite{DBLP:conf/sibgrapi/LarocaSELM22} find that even the LP datasets from the same country also have data bias. However, recent deep learning based LP detectors are generally trained on manually annotated datasets with limited scenarios. Thus, the performance and robustness of LP detectors are restricted by the training data. On the country, the foundation models are generally trained on broad data. Benefiting from the huge training data, the foundation models can easily adapted into wide-range downstream tasks. It means vision foundation model (\emph{e.g.} SAM) is more suitable for zero-shot or few-shot transfer \cite{DBLP:journals/corr/abs-2306-02245, DBLP:journals/corr/abs-2304-06790} to downstream tasks. Therefore, SAM \cite{SAM} is robust to resist the influence from data. To alleviate the influence from biased data and boost the generalization ability of LP detectors, we attempt to customize the Segment Anything Model (SAM) for LP detection task and proposed \emph{\textbf{the first}} LP detector based on vision foundation model, named SamLP. 

First, the SAM \cite{SAM} with ViT-B \cite{ViT} has 168M parameters and the SAM \cite{SAM} with ViT-H \cite{ViT} has 632M parameters which need massive computing resources. Due to the limitation of GPU memory, we can not directly fine tuning the SAM \cite{SAM}. Therefore, to customize the SAM for LP detection task, we introduce the Prameter-Efficient Fine-Tuning (PEFT) \cite{DBLP:journals/natmi/DingQYWYSHCCCYZWLZCLTLS23} into the transfer of SAM. Specifically, we proposed a low-rank adaptation (LoRA) \cite{LoRA} tuning strategy in our proposed SamLP to adapt the SAM \cite{SAM} for LP detection datasets. The introducing of LoRA tuning \cite{LoRA} successfully makes the proposed SamLP focus on vehicle LPs rather general objects and segment the mask of LP regions accurately. 

Meanwhile, a special characteristic of SAM \cite{SAM} is that it introduces the position prompts (\emph{e.g.} points, bounding boxes, and masks) in mask prediction task to generate accurate and expected masks. The proposed LoRA \cite{LoRA} tuning method directly fine-tunes the learning of features, but it does not utilize the most distinctive ability of SAM \cite{SAM} (\emph{i.e.} position prompt). Thus, after LoRA \cite{LoRA} tuning, the promptable segmentation ability of SAM \cite{SAM} is suppressed. To alleviate the harmful effect brought by fine-tuning, we design a promptable training strategy to maintain the promptable segmentation ability for LP detection. Under this condition, the proposed SamLP has the potential for subsequent refinement guided by prompts to elaborate the LP detection performance. 


In addition, the reason why we introduce SAM \cite{SAM} to LP detection is that we attempt to use the zero-shot or few-shot transfer ability of SAM \cite{SAM} to alleviate the influence from intrinsic differences and bias about LPs. Thus, we explore the generalization ability of proposed SamLP on several LP detection datasets. The experimental results show that the proposed SamLP only needs few training data for LoRA tuning and it achieves competitive results compared to recent detectors. Simultaneously, the proposed SamLP has better zero-shot transfer ability and the experiments show that the SamLP trained on LPs from a specific country easily obtains promising detection performance on LPs from other countries.

In summary, the main contributions of this paper are:
\begin{enumerate}
  \item Propose \emph{\textbf{the first}} LP detector based on vision foundation model, \emph{i.e.} SamLP. The proposed SamLP is a customized Segment Anything Model (SAM) \cite{SAM} for LP detection. To adapt the original SAM \cite{SAM} for expected task, we propose a LoRA \cite{LoRA} fine-tuning strategy to fine-tune the original SAM \cite{SAM} on LP detection task. And the proposed fine-tuning strategy make SamLP achieve accurate LP detection successfully.
  \item Propose a promptable fine-tuning step for promptable LP detection. The most distinct characteristic of SAM \cite{SAM} is that it achieves promptable segmentation of objects, but this characteristic is limited by LoRA fine-tuning strategy. To fully exploit the ability of SAM \cite{SAM}, we design a promptable training strategy to alleviate the influence on promptable segmentation. And this strategy makes the refinement with prompt possible, increasing the application of SamLP. 
  \item The zero-shot and few-shot transfer ability of proposed SamLP are explored. The experimental results show that only 3\% training data already achieves promising LP detection accuracy under our proposed LoRA fine-tuning strategy. Moreover, the SamLP also has excellent zero-shot transfer ability. SamLP trained on the LPs from a specific country can easily tackle the detection of LPs from other countries.
\end{enumerate}

The rest of this paper is organized as follows. Section \ref{sec:related} breifly reviews the related works. Section \ref{sec:method} describes the details of proposed SamLP. Section \ref{sec:exp} demonstrates the experimental results on LP detection datasets and analyzes the few-shot and zero-shot transfer ability of SamLP. Finally, we summarize the whole work in Section \ref{sec:con}.

\section{Related Work}\label{sec:related}
In this section, we first introduce the recent researches about vision foundation models in Section \ref{sec:related:FM}, especially SAM \cite{SAM} and its applications. And Section \ref{sec:related:PEFT} introduces recent Parameter-Efficient Fine-Tuning methods for foundation model. Recent LP detectors are reviewed in Section \ref{sec:related:LP}.

\vspace{-8pt}
\subsection{Vision Foundation Models}\label{sec:related:FM}
The foundation models \cite{ChatGPT, GPT4, DBLP:conf/naacl/DevlinCLT19} in NLP demonstrates that foundation models have unparalleled power to tackle challenging tasks. The development of foundation models in NLP promotes the research of foundation model in CV. Some beneficial attempts have emerged for vision tasks. Kirillov \emph{et al.} propose the first vision foundation model, Segment Anything Model \cite{SAM}, for promptable segmentation task. SAM predicts the possible masks at region marked by prompts. Another popular vision foundation model is SegGPT \cite{SegGPT} which introduces the in-context learning \cite{GPT3} into segmentation task. SegGPT \cite{SegGPT} segments interested objects according to the one-shot or few-shot guidance. 

The vision foundation models have been adapted into various vision task. In medical image segmentation task, Zhang \emph{et al.} \cite{DBLP:journals/corr/abs-2304-13785} utilize LoRA \cite{LoRA} to fine-tune the SAM into medical image segmentation. Wu \emph{et al.} introduce Adapter \cite{Adapter} into fine-tuning of SAM \cite{SAM} for medical image. As for object tracking or detection, Tracking Anything \cite{DBLP:journals/corr/abs-2304-11968} uses SAM \cite{SAM} extract object proposals and sends these proposals to XMem \cite{DBLP:conf/eccv/ChengS22} model for tracking. SAM-Adapter \cite{DBLP:journals/corr/abs-2304-09148} fine-tunes the SAM for camouflage object detection. SAM3D \cite{DBLP:journals/corr/abs-2306-02245} directly utilizes SAM \cite{SAM} to segment the objects in bird's eye view images and locates the objects in point clouds from segmentation results. As for image editing, Inpaint Anything \cite{DBLP:journals/corr/abs-2304-06790} applies SAM \cite{SAM} for the segmentation of editing region and then uses an inpainting model to achieve the editing. InpaintNeRF360 \cite{DBLP:journals/corr/abs-2305-15094} also uses SAM \cite{SAM} to segment the editing region in data for NeRF training. Zhang \emph{et al.} \cite{DBLP:journals/corr/abs-2305-03048} propose a training-free one-shot method based on SAM \cite{SAM} to find the given objects in images and guide the generation of stable diffusion \cite{DBLP:conf/cvpr/RombachBLEO22}. 

There are some researches to increase the segmentation performance and efficiency of SAM \cite{SAM}. SAMAug \cite{DBLP:journals/corr/abs-2307-01187} attempt to refine the point prompt selection strategy for better performance. FastSAM \cite{DBLP:journals/corr/abs-2306-12156} and MobileSAM \cite{DBLP:journals/corr/abs-2306-14289} focus on increasing the inference speed to achieve the real-time application of SAM \cite{SAM}.

\vspace{-8pt}
\subsection{Parameter-Efficient Fine-Tuning (PEFT)}\label{sec:related:PEFT}
Due to the limitation of computation resource, fine-tuning the full foundation model seems impractical for most of researchers. Thus, the conception of Parameter-Efficient Fine-Tuning (PETF) is proposed for adapting the foundation model for expected tasks. It aims to fine-tuning the foundation model by using only a part of parameters rather than whole model parameters. There are three types mainstream PETF strategy, \emph{i.e.} adapter tuning, prompt tuning, and LoRA tuning. Adapter tuning \cite{DBLP:conf/icml/HoulsbyGJMLGAG19} is first proposed in NLP to fine-tuning the BERT \cite{DBLP:conf/naacl/DevlinCLT19} model on diverse text classification tasks. AdapterFusion \cite{DBLP:conf/eacl/PfeifferKRCG21} and K-Adapter \cite{DBLP:conf/acl/WangTDWHJCJZ21} are proposed to alleviate the catastrophic forgetting in adapter fine-tuning. Meanwhile, AdapterDrop \cite{DBLP:conf/emnlp/RuckleGGBP0G21} aims to increase the efficiency of training and inference. Adapter tuning \cite{DBLP:conf/icml/HoulsbyGJMLGAG19} is also introduced into CV. ViT-Adapter \cite{DBLP:conf/iclr/ChenDWHLDQ23} uses an adapter to introduce the image-related inductive biases into ViT \cite{ViT}, making it suitable for downstream tasks. Prompt tuning \cite{DBLP:journals/csur/LiuYFJHN23} is another popular PEFT method. GPT-3 \cite{GPT3} first proposes in-context learning in NLP which is the embryo of prompt tuning \cite{DBLP:journals/csur/LiuYFJHN23}. The prompts are task-specific learnable vectors in recent works \cite{DBLP:conf/acl/LiL20, DBLP:conf/emnlp/LesterAC21, DBLP:journals/corr/abs-2110-07602} which only fine-tune the parameters of prompts rather than the full model. Jia \emph{et al.} \cite{VPT} first introduce prompt into ViT model and propose Visual Prompt Tuning (VPT). And other prompt tuning methods \cite{DBLP:conf/eccv/JuHZZX22, Khattak_2023_CVPR, DBLP:conf/nips/WangYR0L22, Sohn_2023_CVPR} focus on the different image based, video based, and point cloud based vision tasks. LoRA tuning \cite{LoRA} is also a well-known PEFT method. LoRA tuning \cite{LoRA} fine-tunes trainable low-rank decomposition matrices for each layer in transformer, reducing the the number of trainable parameters significantly. In this paper, we introduce LoRA tuning \cite{LoRA} into SAM \cite{SAM} to transfer the original SAM \cite{SAM} into LP detection task.

\vspace{-8pt}
\subsection{LP Detection}\label{sec:related:LP}
Recent LP detectors are mainly inherited from general object detectors. For example, early LP detectors \cite{DBLP:journals/soco/RafiquePJ18, DBLP:conf/bmvc/DongHLLZ17, TE2E} are two-stage detectors \cite{FasterR-CNN} for LP detection. To increase the efficiency, one-stage detectors \cite{YOLO, YOLOv2, YOLOv3, YOLOv4} are widely used in LP detection \cite{DBLP:conf/avss/HsuACS17, UFPR-ALPR, DBLP:journals/corr/abs-1909-01754, DBLP:journals/tii/0001WHHSC021, DBLP:journals/ict-express/JamtshoRW21, shahidi2022deep}. In addition, recent works focus on the increasing of accuracy and robustness of LP detectors. For example, Silva and Jung \cite{DBLP:conf/eccv/SilvaJ18, DBLP:journals/tits/SilvaJ22} propose the Warped Planar Object Detection Network (WPOD-NET) and Improved Warped Planar Object Detection Network (IWPOD-NET) to detect LPs under unconstrained scenarios. Similar to WPOD-NET \cite{DBLP:conf/eccv/SilvaJ18}, Wang \emph{et al.} \cite{DBLP:journals/tits/WangBZC22} and Fan \emph{et al.} \cite{fan2022improving} also detect four corners of LPs for accurate LP detection. Chen \emph{et al.} introduce the relationship between vehicle and plate to improve the robustness of LP detectors. Lee \emph{et al.} \cite{DBLP:journals/tits/LeeJKJP22} and Ding \emph{et al.} \cite{SCCA, CLPD} utilize contrastive learning to distinguish the LPs from similar scene texts. Chen and Wang \cite{DBLP:journals/tits/ChenW22} treat LP detection as a segmentation task, and we also use the segmentation ability of SAM \cite{SAM} in this paper to segment and detect LPs. As for video LP detection task, flow-guided methods \cite{Zhangcong, DBLP:conf/mir/LuYW21, LSVLP} are still  the main stream, but the efficiency of these methods are quite low due to the complex computation of optical flow. 

To our best knowledge, there is no work about transferring vision foundation model (\emph{i.e.} SAM \cite{SAM}) into LP detection task, and our proposed SamLP is \textbf{\emph{the first}} LP detector based on vision foundation model.

\vspace{-5pt}
\section{Method}\label{sec:method}
We describe the details of our proposed SamLP in this section. First, we briefly review the fundamental details of SAM \cite{SAM} in Section \ref{sec:method:SAM}. Section \ref{sec:method:pipe} first expounds the whole pipeline of our proposed SamLP. And the LoRA fine-tuning strategy in our proposed SamLP is described in Section \ref{sec:method:LoRA}. Section \ref{sec:method:prompt} illustrates the fine-tuning strategy for promptable segmentation. Finally, Section \ref{sec:method:details} describes some details about proposed SamLP.

\vspace{-8pt}
\subsection{Revisit of Segment Anything Model (SAM)}\label{sec:method:SAM}

\begin{figure}[h]
	\centering
	\includegraphics[width=1.0\linewidth]{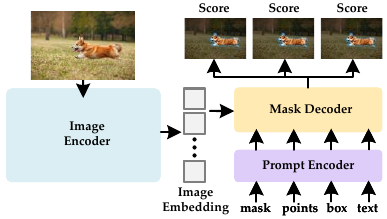}
	\caption{The architecture of Segment Anything Model (SAM) \cite{SAM}. }
	\label{fig:sam}
\end{figure}

SAM \cite{SAM} tackles a novel promptable segmentation task, which means given a prompt (\emph{e.g.} point, box, mask) to SAM \cite{SAM} and it returns a segmentation mask at position of this prompt. SAM \cite{SAM} is fully pre-trained by 1 billion masks from 11 million images (SA-1B dataset \cite{SAM}), which enables SAM \cite{SAM} to gain excellent generalization capacity on segmentation task. As shown in Fig. \ref{fig:sam}, it mainly contains three parts, \emph{i.e.} image encoder $\mathbf{Enc}_{I}$, prompt encoder $\mathbf{Enc}_{P}$, and mask decoder $\mathbf{Dec}_{M}$. During the promptable segmentation, an image $\boldsymbol{I}$ and a set of prompts $\boldsymbol{P}$ (\emph{e.g.} points on foregrounds or backgrounds, bounding boxes, or masks) are fed into SAM. $\mathbf{Enc}_{I}$ extracts features from image $\boldsymbol{I}$ and $\mathbf{Enc}_{P}$ encodes tokens from given prompts $\boldsymbol{P}$, which is illustrated in Eq. \ref{eq:SAM_Enc}:

\begin{equation}
	\label{eq:SAM_Enc}
	\boldsymbol{F}_{I} = \mathbf{Enc}_{I} (\boldsymbol{I}), \enspace \boldsymbol{T}_{P} = \mathbf{Enc}_{P}(\boldsymbol{P}),
\end{equation} 
where $\boldsymbol{F}_{I}$ is the image feature and $\boldsymbol{T}_{P}$ is the prompt token. After that, $\boldsymbol{F}_{I}$ and $\boldsymbol{T}_{P}$ are input to $\mathbf{Dec}_{M}$ for mask prediction. In $\mathbf{Dec}_{M}$, the prompt token $\boldsymbol{T}_{P}$ are concatenated with learnable token $\boldsymbol{T}_{M}$ to generate mask tokens for mask prediction. And the interaction between mask tokens and feature $\boldsymbol{F}_{I}$ are responsible for the final segmentation. The processing in $\mathbf{Dec}_{M}$ is shown as:
\begin{equation}
	\label{eq:SAM_Dec}
	\boldsymbol{\hat{S}}, \boldsymbol{Score}, \boldsymbol{logits} = \mathbf{Dec}_{M} (\boldsymbol{F}_{I}, \mathrm{Concat}(\boldsymbol{T}_{M}, \boldsymbol{T}_{P})),
\end{equation} 
where $\boldsymbol{\hat{S}} = \{\boldsymbol{\hat{S}}_1,\boldsymbol{\hat{S}}_2,\boldsymbol{\hat{S}}_3\}$  is the output masks from SAM \cite{SAM} which contains three levels predictions. The $\boldsymbol{Score}= \{\boldsymbol{Score}_1,\boldsymbol{Score}_2,\boldsymbol{Score}_3\}$ is the corresponding Intersection over Union (IoU) scores to $\boldsymbol{\hat{S}}$, and $\boldsymbol{logits}=\{\boldsymbol{logits}_1,\boldsymbol{logits}_2,\boldsymbol{logits}_3\}$ is the feature logits about $\boldsymbol{\hat{S}}$. $\mathrm{Concat}(\cdot, \cdot )$ is the concatenation operation from tokens.


\begin{figure*}[t]
	\includegraphics[width=0.99\linewidth]{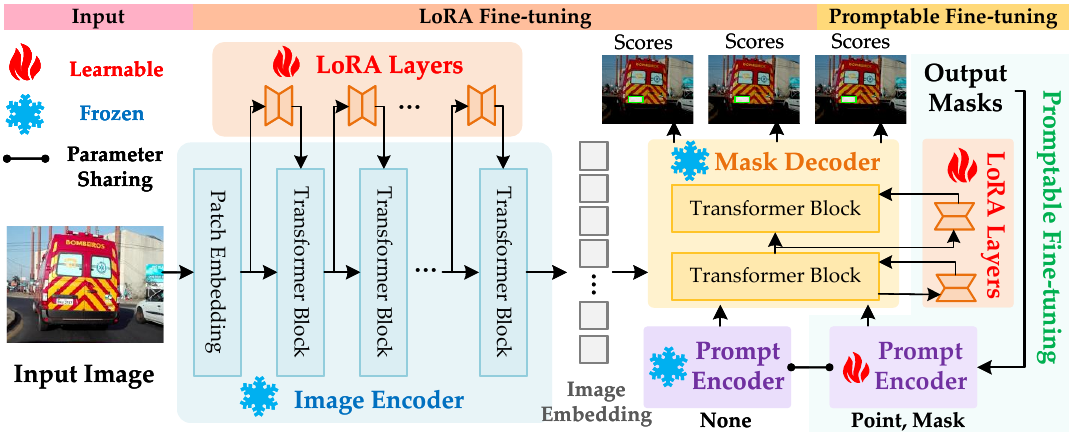}
	\caption{The pipeline of the proposed SamLP. }
	\label{fig:pipline}
\end{figure*}

\subsection{Pipeline}\label{sec:method:pipe}
The whole pipeline of proposed SamLP is shown in Fig. \ref{fig:pipline}. Given an image $\boldsymbol{I} \in \mathbb{R}^{H \times W \times 3}$ whose spatial dimension is $H \times W$ and channel number is $3$ (\emph{i.e.} RGB channels), the aim of proposed SamLP is to predict the segmentation mask $\boldsymbol{\hat{S}} \in \mathbb{R}^{H \times W \times 1}$ of LPs in $\boldsymbol{I}$ where every pixel belongs to a defined category in $\boldsymbol{Y}=\{y_0,y_1\}$. $y_0$ is the background class and $y_1$ denotes the foreground class (\emph{i.e.} LPs). In training, the proposed SamLP should make $\boldsymbol{\hat{S}}$ as close as to $\boldsymbol{S}$. However, due to the limitation of computation resources, fully fine-tuning of SAM \cite{SAM} is impossible. Thus, the proposed SamLP use a parameter-efficient fine-tuning method to transfer original SAM \cite{SAM} into LP detection task. In order to prevent excellent segmentation capacity of SAM \cite{SAM}, the parameters of original SAM \cite{SAM} are froze during training. Specifically, there are two steps for fine-tuning, \emph{i.e.} LoRA fine-tuning and promptable fine-tuning. 

First, the LoRA fine-tuning strategy inserts several LoRA layers into transformer blocks from image encoder and mask decoder of SAM \cite{SAM}. The introduced LoRA fine-tuning adapts image encoder for the extraction of LP-related features and makes mask decoder generate LP masks. In the LoRA fine-tuning step, the position prior of LP is unknown as the input is only a single image $\boldsymbol{I}$, so that there are no information for prompt encoder and the input prompt for LoRA fine-tuning is $\mathrm{None}$. This means the prompt encoder is not adapted into LP detection task. Therefore, to maintain the promptable segmentation capacity of SAM \cite{SAM}, we design the second step, \emph{i.e.} promptable fine-tuning. In promptable fine-tuning, the previous segmentation results of LPs from LoRA fine-tuning are treated as prior prompt to guide the training of promptable segmentation. The full image encoder with its LoRA layers are frozen and only prompt encoder and LoRA layers in mask decoder are trained to avoid the catastrophic forgetting of LP features in image embedding and accelerate the training in promptable fine-tuning. The details of LoRA fine-tuning and promptable fine-tuning are described in the following text.
\vspace{-8pt}
\subsection{LoRA Fine-tuning}\label{sec:method:LoRA}

\begin{figure}[h]
	\centering
	\includegraphics[width=0.6\linewidth]{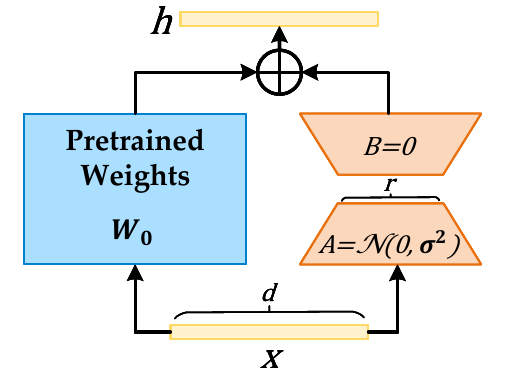}
	\caption{The detail architecture of LoRA layers.}
	\label{fig:lora}
\end{figure}

Low-Rank Adaptation (LoRA) \cite{LoRA} is a parameter-efficient fine-tuning for large foundation model. Aghajanyan \emph{et al.} \cite{DBLP:conf/acl/AghajanyanGZ20} find that the learning ability of a pre-trained language model is still efficient and effective when the model is projected to a low-rank subspace. This means the pre-trained models have a low intrinsic dimension and the updating of parameters also has a low intrinsic rank during the adaptation of model. This characteristic also appears at vision foundation models.

As shown in Fig. \ref{fig:lora}, the pre-trained parameters of foundation model is $\boldsymbol{W_0} \in \mathbb{R}^{d \times k}$ where $d$ is the input feature dimension and $k$ is the output feature dimension. The optimization of $\boldsymbol{W_0}$ can be represented as Eq. \ref{eq:lora}:

\begin{equation}
	\label{eq:lora}
	\begin{split}
		\boldsymbol{\hat{W_0}} & = \boldsymbol{W_0} + \Delta\boldsymbol{W} \\
		& = \boldsymbol{W_0} + \boldsymbol{BA}, \\
	\end{split}
\end{equation} 
where $\Delta\boldsymbol{W}$ is the updated value for model parameters during adaptation, and it is decomposed into two matrices, \emph{i.e.} $\boldsymbol{B} \in \mathbb{R}^{d \times r}$ and $\boldsymbol{A} \in \mathbb{R}^{r \times k}$. $\boldsymbol{B}$ and $\boldsymbol{A}$ are the low rank decomposition matrices in LoRA \cite{LoRA} and its low rank size $r \ll \mathrm{min}(d,k)$. In LoRA fine-tuning, the parameter $\boldsymbol{W_0}$ from SAM \cite{SAM} is frozen and their updating gradient will not be calculated. Only parameters in low rank decomposition matrices, \emph{i.e.} $\boldsymbol{B}$ and $\boldsymbol{A}$, are trainable, which reduce the computation burden in training significantly. 

The input feature $\boldsymbol{x} \in \mathbb{R}^{n \times d}$ ($n$ is the sequence length of flattened input feature $\boldsymbol{x}$) is multiple with both $\boldsymbol{W_0}$ and $\Delta\boldsymbol{W}$ simultaneously, and then their outputs are summed by element-wise as output feature $\boldsymbol{h} \in \mathbb{R}^{n \times k}$:
\begin{equation}
	\label{eq:lora_h}
	\begin{split}
		\boldsymbol{h} & = \boldsymbol{W_0}\boldsymbol{x} + \Delta\boldsymbol{W}\boldsymbol{x} \\
		& = \boldsymbol{W_0}\boldsymbol{x} + \boldsymbol{BA}\boldsymbol{x} \enspace . \\
	\end{split}
\end{equation} 

\begin{figure}[h]
	\centering
	\includegraphics[width=0.8\linewidth]{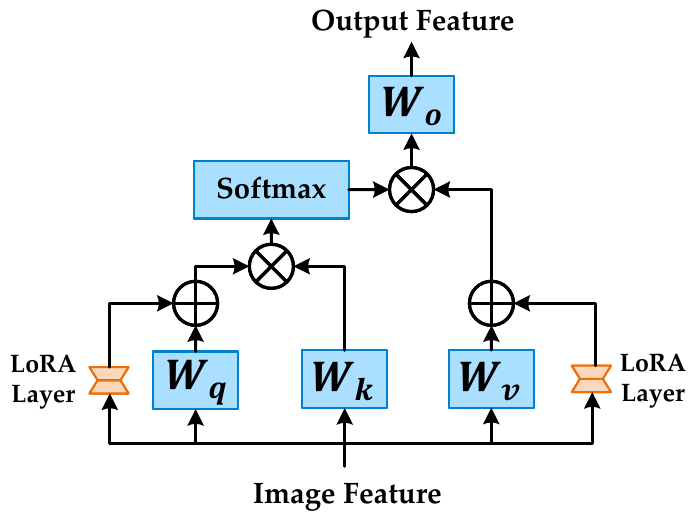}
	\caption{The injection of LoRA layers in transformer blocks.}
	\label{fig:inject}
\end{figure}

In practice, we embed LoRA layers in image encoder and mask decoder to build the SAM \cite{SAM} with LoRA fine-tuning (denote as \textbf{SamLP} in the following content), and there is no injection in prompt encoder. Only parameters in LoRA layers are trainable, and the original parameters in image encoder, prompt encoder, and mask decoder are frozen. This means the LoRA fine-tuning step only fine-tunes the image feature extraction and mask segmentation of SAM \cite{SAM}, instead of the promptable segmentation which is the most distinct capacity of SAM \cite{SAM}. The LoRA can be inject into any weights matrices (\emph{i.e.} model parameters) in neural network. Referring to recent works about LoRA \cite{LoRA, DBLP:journals/corr/abs-2304-13785}, the LoRA layers are injected into the query weights $\boldsymbol{W_q}$ and value weights in every self-attention module or cross-attention module in image encoder and mask decoder, which is demonstrated in Fig. \ref{fig:inject}. For all LoRA layers in image encoder and mask decoder, the low-rank decomposition matrix $\boldsymbol{A}$ is initialized by a random Gaussian distribution $\mathcal{N}(0,\sigma^2)$ ($\sigma=5$ according to \cite{DBLP:journals/corr/abs-2304-13785}), and the initial value of matrix $\boldsymbol{B}$ is zero matrix.

In LoRA fine-tuning step, the input image $\boldsymbol{I}$ is fed into the image encoder with LoRA layers to obtain image embedding. And then, the input prompts $\boldsymbol{P}$ is $\mathrm{None}$ which means there is no point, box, or mask input to prompt encoder. After that, the prompt embedding of $\mathrm{None}$ is combined with mask tokens and then mask decoder retrievals high response regions in image embedding from image encoder to generate binary segmentation masks. The whole procedure described above is illustrated as Eq. \ref{eq:SamLP:LoRA}:
\begin{equation}
	\label{eq:SamLP:LoRA}
	\begin{split}
	\boldsymbol{F}_{I} & = \mathbf{Enc}_{I} (\boldsymbol{I}), \\ 
	\boldsymbol{T}_{P} & = \mathbf{Enc}_{P}(\mathrm{None}),  \\
	\boldsymbol{\hat{S}}, \boldsymbol{Score}, \boldsymbol{logits} & = \mathbf{Dec}_{M} (\boldsymbol{F}_{I}, \mathrm{Concat}(\boldsymbol{T}_{M}, \boldsymbol{T}_{P})),
	\end{split}
\end{equation} 
where $\mathbf{Enc}_{I}$ and $\mathbf{Dec}_{M}$ are the image encoder with LoRA layers and mask decoder with LoRA layers respectively, and $\mathbf{Enc}_{P}$ is the prompt encoder without LoRA layers. The prediction masks $\boldsymbol{\hat{S}}$ contains three masks with different levels defined in SAM \cite{SAM}. The corresponding IoU scores and logits of predicted masks are $\boldsymbol{Score}$ and $\boldsymbol{logits}$ respectively. And all these there masks $\boldsymbol{\hat{S}}_i \in \mathbb{R}^{H \times W} \enspace (i=1,2,3)$ are supervised by ground truth $\boldsymbol{S} \in \mathbb{R}^{H \times W}$ in LoRA fine-tuning. The loss function $\mathcal{L}$ is shown as follow:
\begin{equation}
	\label{eq:loss:LoRA}
	\begin{split}
		\mathcal{L} &= \frac{1}{3} \Sigma_{i=1}^3 \mathrm{Dice}(\boldsymbol{\hat{S}}_i, \boldsymbol{S}) \\
	\end{split}
\end{equation} 
where $ \mathrm{Dice}(\cdot, \cdot)$ is Dice loss \cite{Dice} which is a common loss for segmentation tasks. 



\vspace{-10pt}
\subsection{Promptable Fine-tuning}\label{sec:method:prompt}
\vspace{-10pt}
 \begin{algorithm}[h]
	\label{alg:prompt}
	\KwInput{input image $\boldsymbol{I}$, initial prompt $\boldsymbol{P}$, SamLP, ground truth $\boldsymbol{S}$, iteration times $\mathrm{Num}$}
	\KwResult {SamLP with promptable fine-tuning (\emph{i.e.} SamLP\_P)}
	$\boldsymbol{P} = \mathrm{None}$\;
	\For{t = 0, 1, $\cdots$, $\mathrm{Num}$}{
		$\boldsymbol{\hat{S}}_i, \boldsymbol{Score}_i, \boldsymbol{logits}_i = \mathrm{SamLP}(\boldsymbol{I}, \boldsymbol{P})$\;
		$idx = \mathop{\arg\max}\limits_{i \in \{1,2,3\}}(\boldsymbol{Score}_i)$\;
		$mask\_prompt = \boldsymbol{logits}_{idx}$\;
		Sample a foreground point $pos$ from false negative region from $\boldsymbol{\hat{S}}_{idx}$ supervised by $\boldsymbol{S}$\;
		Sample a background point $neg$  from false positive region from $\boldsymbol{\hat{S}}_{idx}$ supervised by $\boldsymbol{S}$\;
		$point\_prompt = pos,\  neg$\;
		$\boldsymbol{P} = mask\_prompt,\  point\_prompt$\;
	}
	Calculate loss function $\mathcal{L}$ in Eq. \ref{eq:loss:LoRA}\;
	Optimize $\mathcal{L}$ to update SamLP.
	\caption{Algorithm for promptable fine-tuning.}
\end{algorithm}
The LoRA fine-tuning step has already made SamLP segment LPs effectively, but the LoRA fine-tuning in Section \ref{sec:method:LoRA} only adapts the image encoder and mask decoder to LP detection task rather than prompt encoder. Because the LP detection task do not provide any cues (\emph{e.g.} points, boxes, \emph{etc.}) to localize LPs, the prompt encoder is not fully exploited in LoRA fine-tuning step, causing the mismatching between prompt encoder and mask decoder. Under this condition, the SamLP can not achieve accurate promptable segmentation for LPs. Thus, to maintain promptable segmentation capacity of SAM \cite{SAM}, the prompt encoder also need to be transferred. 


In SAM \cite{SAM}, the prompt encoder mainly embeds the prompts (\emph{i.e.} points, boxes, masks) into embeddings and these prompt embeddings are combined with tokens in mask decoder to inject the information from prompts into mask prediction. Therefore, in promptable fine-tuning step, we need to fine-tune the prompt encoder and mask decoder.

In promptable fine-tuning step, we design an iterative refinement training pipeline to introduce prompt into segmentation. The procedure of promptable fine-tuning step on an image in training is listed in Algorithm \ref{alg:prompt}.

After promptable fine-tuning, the SamLP maintains the promptable segmentation capacity (denoted as \textbf{SamLP\_P} in the following content). However, due to the limitation of data diversity, the promptable fine-tuning step is not always effective, which is discussed in Section \ref{sec:exp:abl}. But the promptable fine-tuning (\emph{i.e.} SamLP\_P) indeed increases the promptable segmentation performance compared to the SamLP which is described in Section \ref{sec:method:LoRA}.  

\vspace{-10pt}
\subsection{Other Details}\label{sec:method:details}
In the training of SamLP, we train the SAM \cite{SAM} by LoRA fine-tuning to get proposed SamLP (described in Section \ref{sec:method:LoRA}), and the LoRA fine-tuning has already empowered SamLP to segment LPs successfully. In addition, if we need the promptable segmentation capacity of SAM \cite{SAM}, we further fine-tune the SamLP by promptable fine-tuning described in Section \ref{sec:method:prompt} to get SamLP\_P.

As for the SamLP, the input image and $\mathrm{None}$ prompt are input to SamLP for both training and inference. As for the SamLP\_P, the input image and prompt (\emph{i.e.} point and mask) are input to the SamLP\_P for training. And in inference, only mask prompt is fed to SamLP\_P to achieve better detection accuracy, which is discussed in Section \ref{sec:exp:abl}. In both LoRA fine-tuning and promptable fine-tuning, three level masks output from SamLP or SamLP\_P are all supervised by annotated label (\emph{i.e.} Eq. \ref{eq:loss:LoRA}) to ensure all masks focus on LPs. 

\vspace{-10pt}
\section{Experiments}\label{sec:exp}
This section mainly demonstrates the experimental results about the proposed SamLP. Section \ref{sec:exp:data} introduces the datasets used in LP detection task. Section \ref{sec:exp:eval} claims the evaluation metrics in quantitative comparison. The implementation details are described in Section \ref{sec:exp:imp}. The detection performance of SamLP is demonstrated in Section \ref{sec:exp:perf}. In addition, the ablation study is explained in Section \ref{sec:exp:abl} to analyze the effect of SamLP.

\vspace{-10pt}
\subsection{Dataset}\label{sec:exp:data}
In this paper, several LP detection datasets are used to evaluate the proposed SamLP. 

\subsubsection{UFPR-ALPR} UFPR-ALPR \cite{UFPR-ALPR} dataset is a high-resolution LP detection and recognition dataset. It contains 1800 images for training, 600 images for validation, and 1800 images for testing. All images in UFPR-ALPR \cite{UFPR-ALPR} have the size of $1,920 \times 1,080$ pixels. In a single image from dataset, only a single plate is contained and annotated. The types of LP in UFPR-ALPR \cite{UFPR-ALPR} are various, including gray LP, red LP, and motorcycle LP.

\subsubsection{CCPD} CCPD \cite{CCPD} dataset includes a huge amount Chinese LPs captured in parking lots. The images in CCPD \cite{CCPD} all have a fixed size of $720 \times 1,160$. There are nine subset in CCPD \cite{CCPD} containing different scenes: CCPD-Base ($200,000$ images), CCPD-DB ($20,000$ images), CCPD-FN ($20,000$ images), CCPD-Blur (more than $20,000$ images), CCPD-Rotate ($10,000$ images), CCPD-Tilt ($10,000$ images), CCPD-Weather ($10,000$ images), CCPD-Challenge ($10,000$ images), and CCPD-NP (more than $3,000$ images). Each image only contains a single LP, and the LP types are diverse. 

\subsubsection{CRPD} CRPD \cite{CRPD} dataset is a Chinese LP dataset which provides more than $3,000$ high-resolution images. The above LP detection datasets only contains a single LP in an image. CRPD \cite{CRPD} provides three subset: CRPD-single, CRPD-double, and CRPD-multi which contain a single LP, two LPs, and more than two LPs respectively. The types of LP are also various, including blue plates, yellow and single-line plates, yellow and double-lines plates, and white plates. 

\subsubsection{AOLP} AOLP \cite{AOLP} contains $2,049$ Taiwan LPs. They are split into three subset according to different scenes: access control (AC) with 681 images, traffic law enforcement (LE) with 757 images, and road patrol (RP) with 611 images. The images have different lighting and weather conditions. 

\vspace{-10pt}
\subsection{Evaluation Metrics}\label{sec:exp:eval}
There are some general evaluation metrics for detection task, illustrating in Eq. \ref{eq:eval}:
\begin{equation}
	\label{eq:eval}
	\begin{aligned}
		&\mathrm{P} = \frac{\mathrm{TP}}{\mathrm{TP}+\mathrm{FP}},\\
		&\mathrm{R} = \frac{\mathrm{TP}}{\mathrm{TP}+\mathrm{FN}}, \\
		&\mathrm{IoU}(\boldsymbol{pred},\boldsymbol{gt}) = \frac{\boldsymbol{pred}\cap \boldsymbol{gt}}{\boldsymbol{pred}\cup \boldsymbol{gt}}, \\
		&\mathrm{F1\mbox{-}score} = \frac{2\times \mathrm{P} \times \mathrm{R}}{\mathrm{P}+\mathrm{R}}, \\
		&\mathrm{AP} = \int_0^1 p(r) dr,
	\end{aligned}
\end{equation}
where $\mathrm{TP}$, $\mathrm{FP}$, and $\mathrm{FN}$ are the number of True Positive, False Positive, and False Negative samples respectively. $\mathrm{P}$, $\mathrm{R}$ represent the Precision, Recall in detection. And $\mathrm{IoU}$ is the Intersection over Union (IoU) between predicted bounding box and ground truth. To consider Precision and Recall simultaneously, $\mathrm{F1\mbox{-}score}$ and $\mathrm{AP}$ are calculated. The $p(r)$ in $\mathrm{AP}$ denotes the Precision-Recall curve.

\vspace{-10pt}
\subsection{Implementation Details}\label{sec:exp:imp}

\subsubsection{Inputs} Following the inputs for SAM \cite{SAM}, the input image is first resized according to the rule that the long side of image is resized to $1,024$ pixels. During the resizing, the aspect ratio of image is fixed to avoid the deformation of image content. After resizing, the resized image is padded to a fixed size of $1,024 \times 1,024$ pixels. Finally, the image is normalized and input to the proposed SamLP. 

\subsubsection{Training Strategy} The SamLP uses the pre-trained SAM \cite{SAM} as the foundation. The parameters from original SAM \cite{SAM} are frozen, and only the LoRA layers that are injected into image encoder and mask decoder are trained on LP datasets. We utilize Stochastic Gradient Descent (SGD) method to optimize the trainable parameters in proposed SamLP. The initial learning rate is $5 \times 10^{-3}$ and the learning rate is updated by multiplying a scale factor related to training iteration. The decreasing learning rate helps SamLP in searching the better LP detection performance. The random seed in training is $0$, and the batch size is $2$ to ensure the training on a single GPU. We train the SamLP with LoRA fine-tuning on $160$ epochs totally and train the SamLP with promptable fine-tuning (\emph{i.e.} SamLP\_P) on $320$ epochs totally.

\subsubsection{Inference Strategy} As for SamLP, the input image and $\mathrm{None}$ prompt are fed into proposed SamLP, and we select the mask with the highest IoU score as finally prediction. As for SamLP\_P, because there is no prior prompt before inference, the input image and $\mathrm{None}$ prompt are first input into SamLP\_P to generate the initial prediction in the first iteration. The output logits with highest IoU score is treated as the mask prompt. And this mask prompt is fed into the next iteration to refine the prediction. The iterative refinement repeats $\mathrm{Num}$ times (in this paper $\mathrm{Num}=1$) to output final LP detection results.

\subsubsection{Environment} The proposed SamLP is trained and tested on a computer with Intel(R) CPU Core(TM) i7-6900K @ 3.4 GHz, 64 GB RAM, and a single NVIDIA RTX3090 GPU. The model is built by Pytorch framework \cite{Pytorch}.


\begin{table*}[t]
	\caption{The comparison results of the proposed SamLP on UFPR-ALPR \cite{UFPR-ALPR}. The proposed SamLP achieves better performance compared to listed LP detection methods. The best results are presented in \textbf{bold}.}
	\centering
	\begin{tabular}{m{0.16\linewidth}|m{0.12\linewidth}<{\centering}|c|m{0.12\linewidth}<{\centering} m{0.12\linewidth}<{\centering} m{0.12\linewidth}<{\centering} m{0.12\linewidth}<{\centering}}
		\toprule[1pt]
		Method & Dataset & Promptable  & P & R & F1-score & AP  \\
		\midrule[1pt]
		FCOS \cite{FCOS} & \multirow{8}*{UFPR-ALPR} & \multirow{8}*{\ding{56}} & 93.6 & 96.7 & 95.1 & 92.9  \\
		YOLOv2 \cite{YOLOv2} & & & 92.7 & 94.7 & 93.7 & 87.8  \\
		YOLOv3 \cite{YOLOv3} &  &  & 94.9 & 97.4 & 96.1 & -  \\
		YOLOv4 \cite{YOLOv4} &  &  & 93.1 & 92.6 & 92.8 & 88.5   \\
		EAST \cite{EAST} & &  & 92.8 & 99.9 & 96.2 & -  \\
		R-FCN \cite{R-FCN} & &  & 94.6 & 99.8 & 94.5 & - \\
		Laroca \emph{et al.} \cite{UFPR-ALPR} & &  & - & 98.3 & - & -  \\
		\midrule[1pt]
		SamLP &  \multirow{2}*{UFPR-ALPR} & \ding{56} & 95.3 & 98.4 & 96.8 & 94.9  \\
		SamLP\_P & & \ding{52} & 95.3 & 98.6 & \textbf{96.9} & \textbf{95.2} \\
		\bottomrule[1pt]
	\end{tabular}
	\label{tab:LP_UFPR}
\end{table*}

\begin{table*}[ht]
	\caption{The comparison results of the proposed SamLP on CCPD \cite{CCPD}. The proposed SamLP achieves better performance compared to listed LP detection methods. The best results are presented in \textbf{bold}.}
	\centering
	\begin{tabular}{m{0.15\linewidth}|m{0.07\linewidth}<{\centering}|c|m{0.05\linewidth}<{\centering} m{0.05\linewidth}<{\centering} m{0.05\linewidth}<{\centering} m{0.05\linewidth}<{\centering}  m{0.05\linewidth}<{\centering} m{0.05\linewidth}<{\centering} m{0.05\linewidth}<{\centering} m{0.07\linewidth}<{\centering}}
		\toprule[1pt]
		Method & Dataset & Promptable  & Base & DB & Blur & FN & Rotate & Tilt & Weather & Challemge  \\
		\midrule[1pt]
		Cascade Classifier \cite{DBLP:journals/tifs/WangL07} & \multirow{10}*{CCPD} & \multirow{10}*{\ding{56}} & 55.4 & 49.2 & - & 52.7 & 0.4 & 0.6  & 51.5 & 27.5 \\ 
		SSD300 \cite{SSD} & & & 99.1 & 89.2  & 87.1 & 84.7 & 95.6 & 94.9  & 83.4 & 93.1  \\ 
		YOLOv2 \cite{YOLOv2} & & & 98.8 & 89.6 & - & 77.3 & 93.3 &  91.8 &  84.2 & 88.6   \\ 
		FCOS \cite{FCOS} & & & 99.3 & - & - & - & - &  - &  - & -   \\ 
		Faster R-CNN \cite{FasterR-CNN} & & & 98.1 & \textbf{92.1} & 81.6 & 83.7 & 91.8 & 89.4  & 81.8 & 83.9  \\ 
		TE2E \cite{TE2E}  & & & 98.5 & 91.7 & - & 83.8 & 95.1 & 94.5 & 83.6 & 93.1  \\ 
		RPnet \cite{CCPD}   & & & 99.3 & 89.5 & - & 85.3 & 84.7 & 93.2 & 84.1 & 92.8  \\ 
		Silva and Jung \cite{DBLP:journals/tits/SilvaJ22}  & & & - & 86.1 & - & 84.3 & 94.8 & 93.0 & 95.7 & 93.4  \\ 
		SCCA \cite{SCCA} & & & 99.6 &- &- &- &- &-  &- &-  \\ 
		CLPD \cite{CLPD} & & & 99.8 & - & - &- &- &-  &- &-  \\ 
		\midrule[1pt]
		SamLP & \multirow{2}*{CCPD} &  \ding{56} & \textbf{99.9} & 91.8 & \textbf{94.3} & 92.6 & 
		98.6 & \textbf{98.8} & 99.8 & 95.4   \\ 
		SamLP\_P & &\ding{52} & \textbf{99.9} & 88.0 & 94.0  & \textbf{93.1} & \textbf{98.8} & \textbf{98.8} & \textbf{99.9} & \textbf{95.9}  \\ 
		\bottomrule[1pt]
	\end{tabular}
	\label{tab:LP_CCPD}
\end{table*}

\begin{table*}[ht]
	\caption{The comparison results (Precision) of the proposed SamLP on CRPD \cite{CRPD}. The proposed SamLP achieves better performance compared to listed LP detection methods. The best results are presented in \textbf{bold}.}
	\centering
	\begin{tabular}{m{0.14\linewidth}|c|c|m{0.08\linewidth}<{\centering} m{0.08\linewidth}<{\centering} m{0.08\linewidth}<{\centering}|m{0.08\linewidth}<{\centering} m{0.08\linewidth}<{\centering} m{0.08\linewidth}<{\centering}}
		\toprule[1pt]
		\multirow{2}*{Method} &  \multirow{2}*{Dataset} &  \multirow{2}*{Promptable} & \multicolumn{3}{c|}{CRPD\_single} & \multicolumn{3}{c}{CRPD\_multi}   \\
		\cmidrule[1pt]{4-9}
		& & & P & R & F1-score & P & R & F1-score  \\
		\midrule[1pt]
		SSD512 \cite{SSD}  &  \multirow{8}*{CRPD} & \multirow{8}*{\ding{56}} & 98.9 & 28.7 & 44.4 & 93.5 & 21.1 & 34.5  \\
		YOLOv3 \cite{YOLOv3} & &  & 73.7 & 59.4 & 65.8 & 66.2 & 61.3 & 63.6 \\
		YOLOv4 \cite{YOLOv4} & &  & 87.3  & 68.4 & 76.7  & 88.9 & 36.8 & 52.0  \\
		Scale-YOLOv4 \cite{ScaleYOLOv4} & & & 90.1 & 72.4 & 80.3 & 91.5 & 75.2 & 72.1 \\
		Faster-RCNN \cite{FasterR-CNN} & & & 81.4 & 71.7 & 76.3 & 69.3 & 75.2 & 72.1  \\
		STELA \cite{STELA} & &  & 83.1 & 73.3 & 77.9 & 77.6 & 82.8 & 80.1 \\
		CRPD \cite{CRPD} & &  & 96.3 & 83.6 & 89.5 & 90.8 &85.0  & 87.7 \\
		HFENet \cite{HFENet} & &  & \textbf{99.0} & 97.0 & 98.0 & 93.2 & 94.9 & 94.1  \\
		\midrule[1pt]
		SamLP &  \multirow{2}*{CRPD} &  \ding{56} & 97.1 & 98.9 & 98.0 & \textbf{94.3} & 96.8 & \textbf{95.5} \\
		SamLP\_P & & \ding{52} & 98.7 & \textbf{99.3} & \textbf{99.0} & 94.0 & \textbf{97.1} & \textbf{95.5} \\
		\bottomrule[1pt]
	\end{tabular}
	\label{tab:LP_CRPD}
\end{table*}

\vspace{-8pt}
\subsection{Performance}\label{sec:exp:perf}

\subsubsection{Detection Performance}
We first test the LP detection performance on several LP detection datasets. In these experiments, the SamLP is trained by the full training data. The experimental results show the detection performance of our proposed SamLP, and verify the effectiveness of the proposed fine-tuning strategy (\emph{i.e.} SamLP\_P). 

\begin{figure}[h]
	\centering
	\subfloat{
		\includegraphics[width=0.49\linewidth]{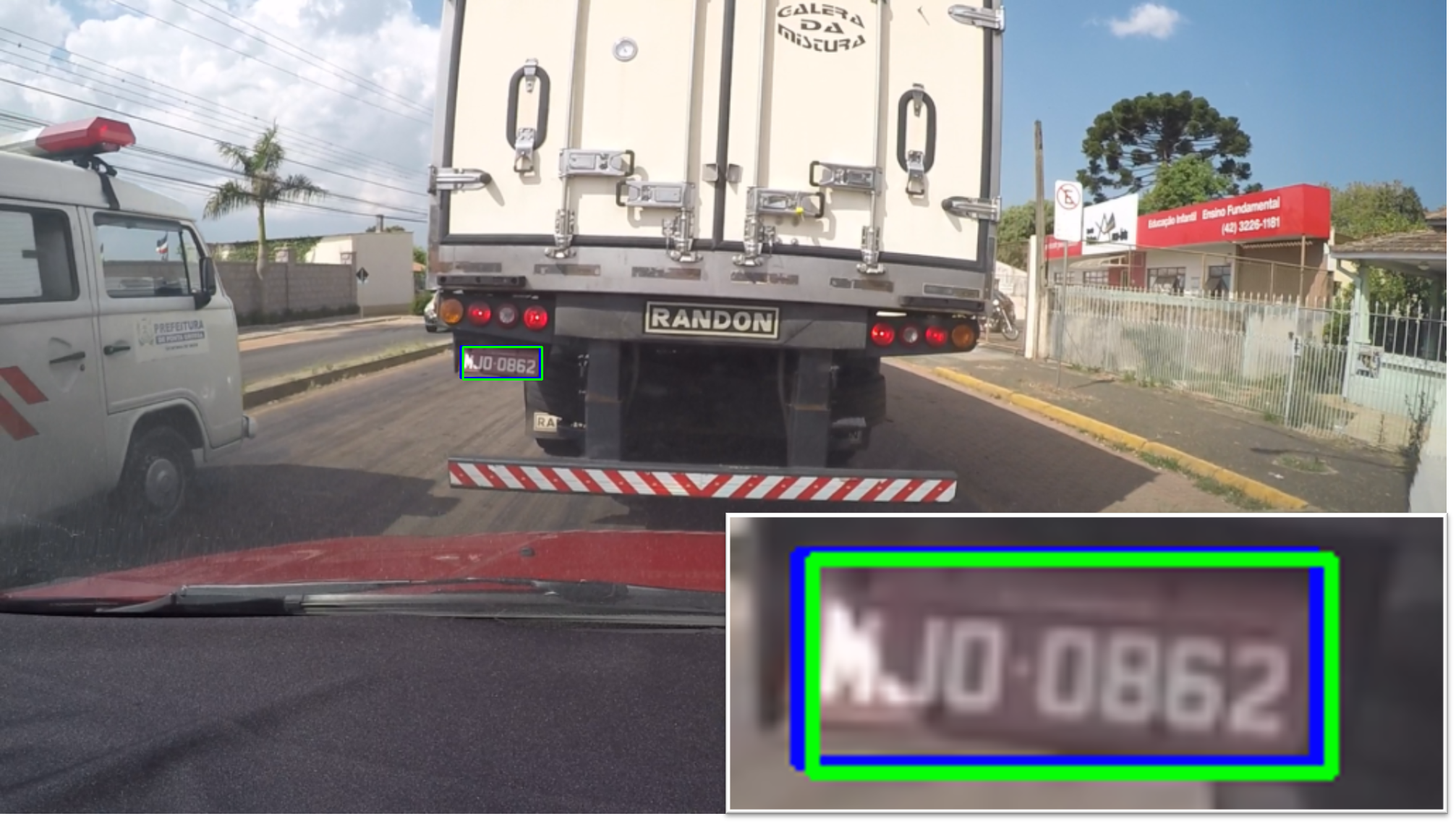}
		\includegraphics[width=0.49\linewidth]{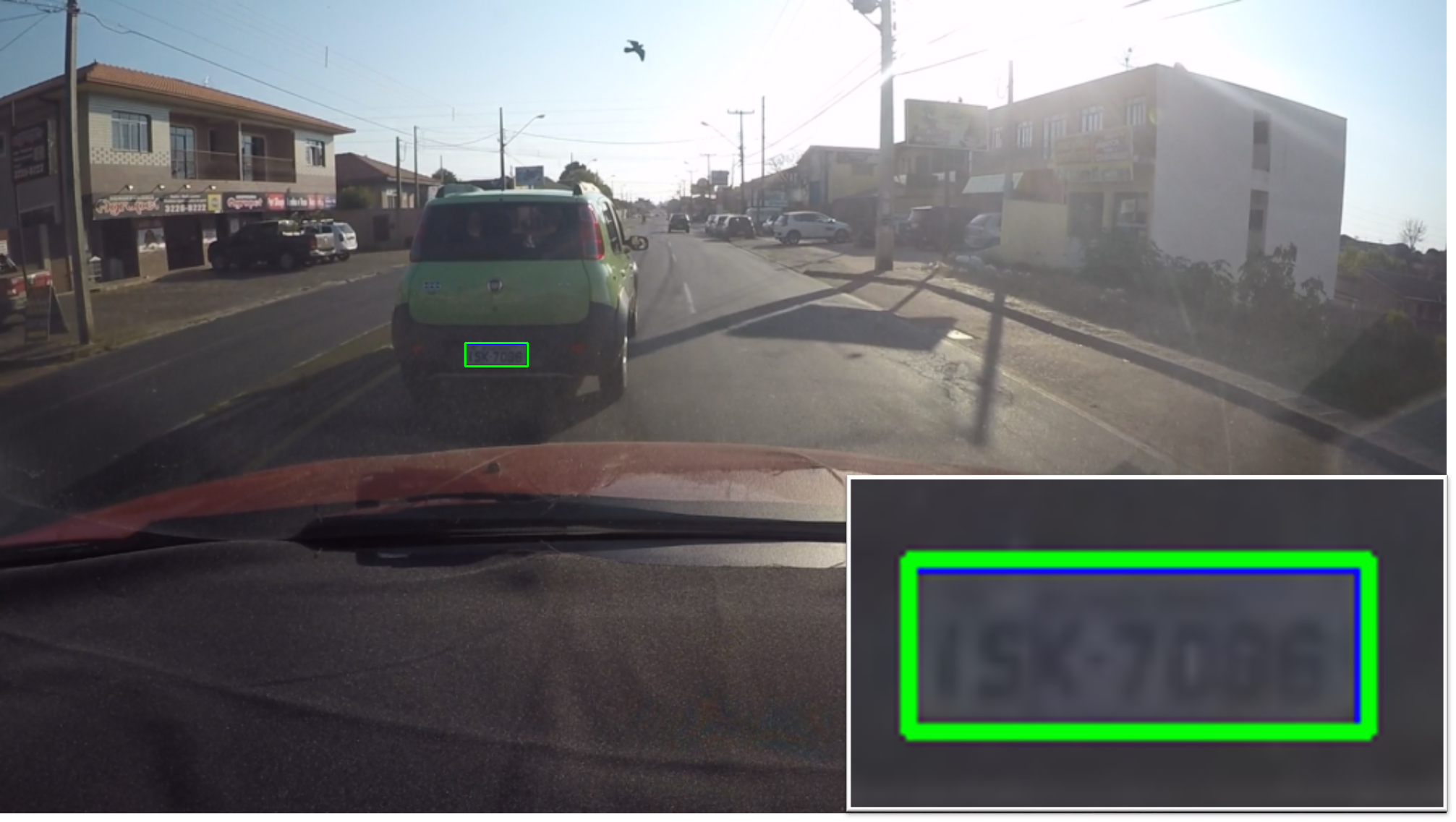}
	}
	\hfil
	\subfloat{
		\includegraphics[width=0.49\linewidth]{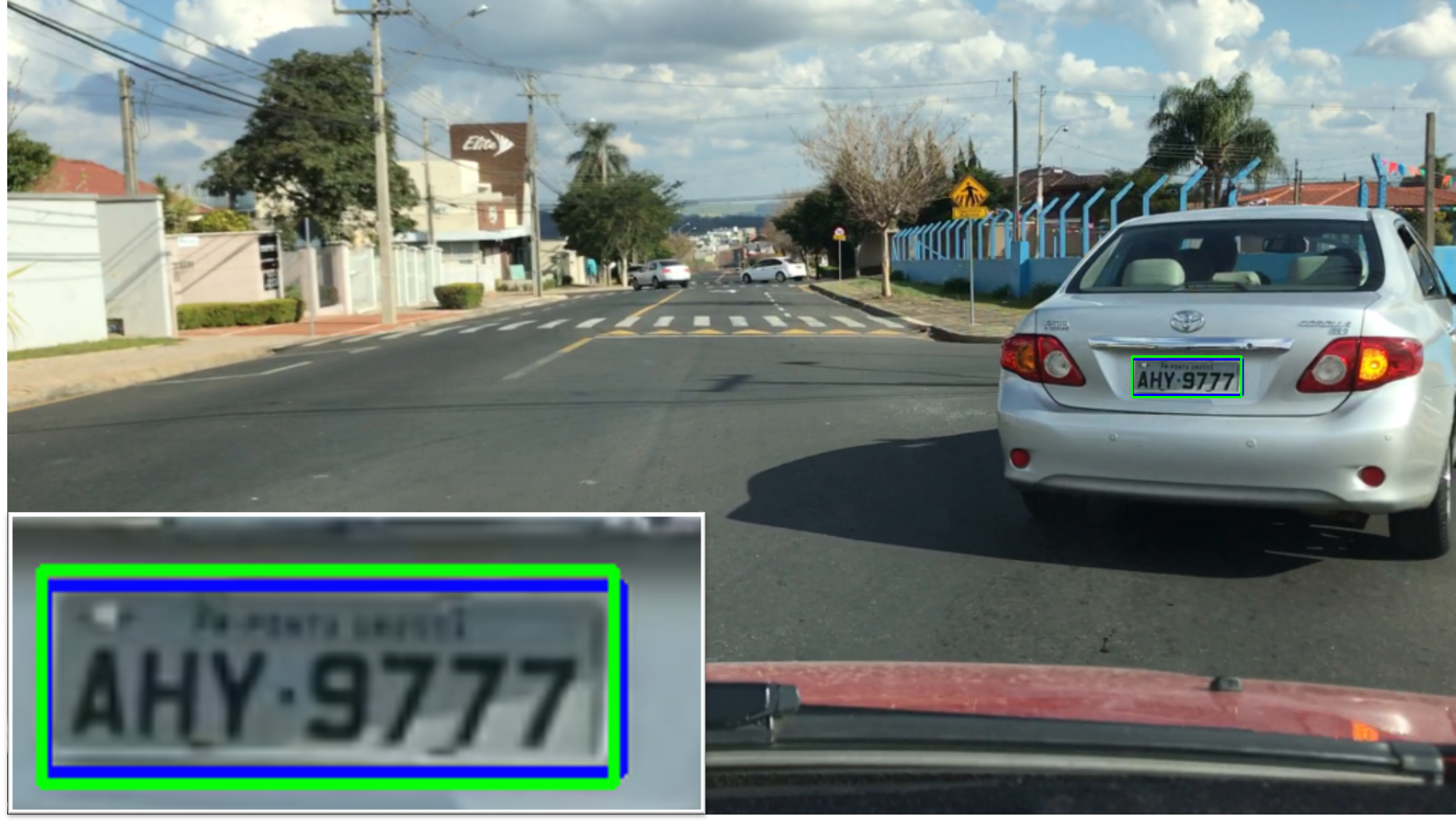}
		\includegraphics[width=0.49\linewidth]{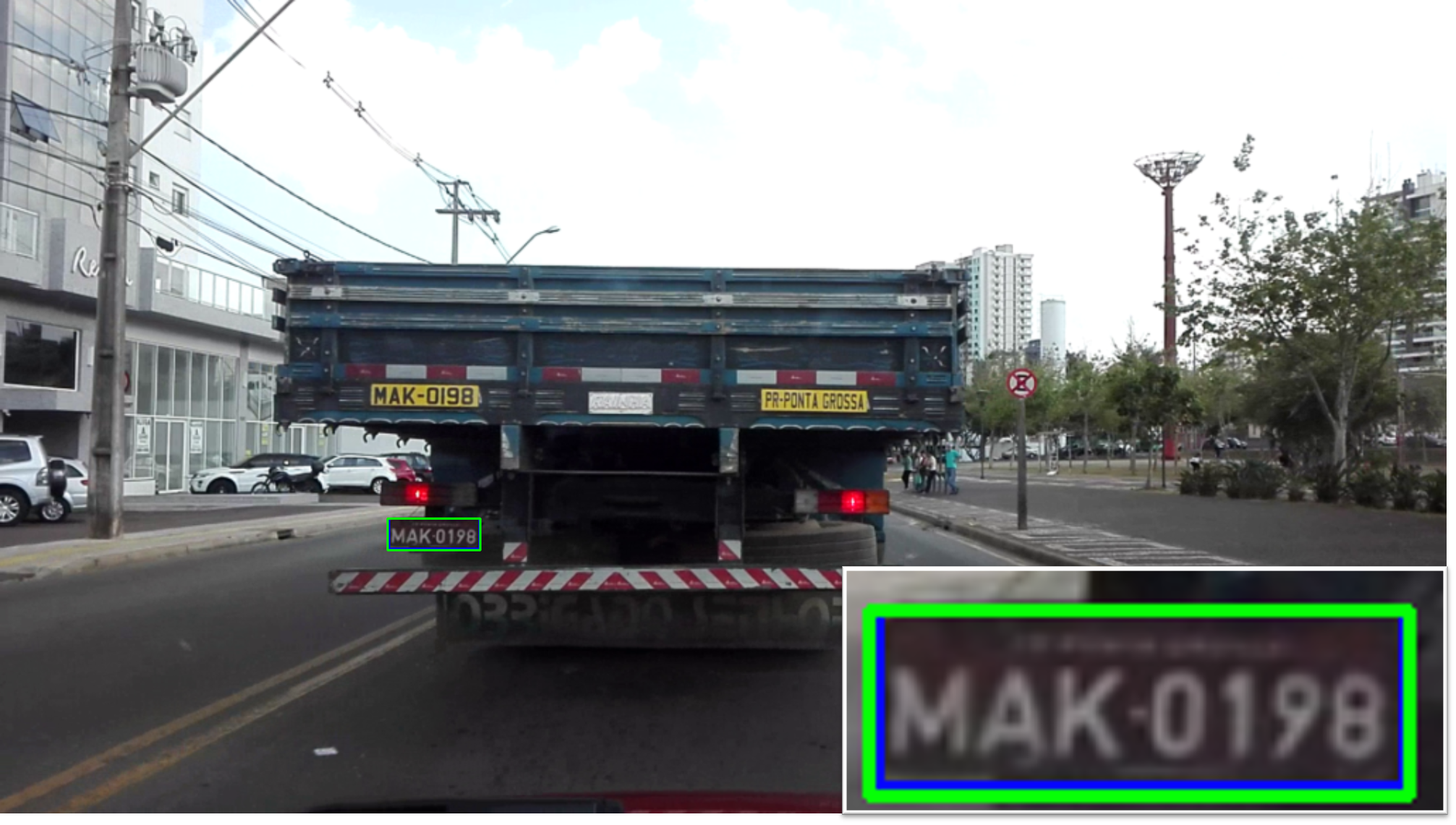}
	}
	\caption{Some detection results of our proposed SamLP on UFPR-ALPR \cite{UFPR-ALPR}. The \textcolor{blue}{blue boxes} are ground truths and the \textcolor{green}{green boxes} are predictions from SamLP.}
	\label{fig:ufpr}
	\vspace{-8pt}
\end{figure}

\begin{figure}[h]
	\centering
	\subfloat{
		\includegraphics[width=0.45\linewidth]{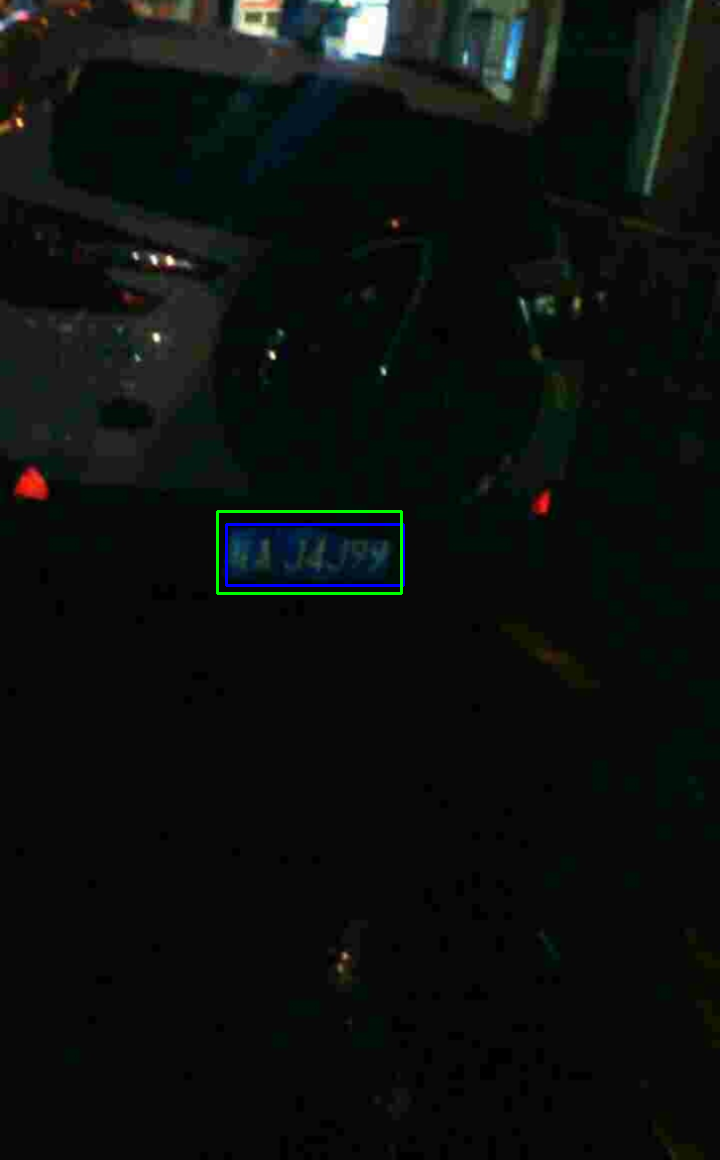}
		\includegraphics[width=0.45\linewidth]{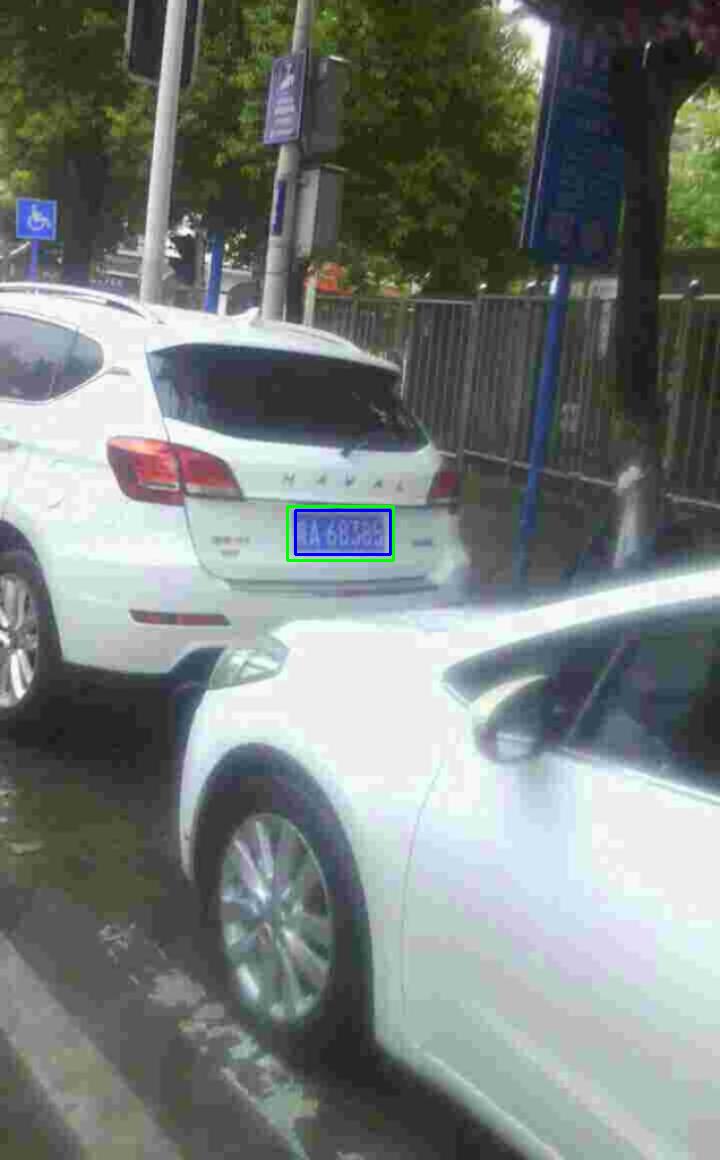}
	}
	\caption{Some detection results of our proposed SamLP on CCPD \cite{CCPD}. The \textcolor{blue}{blue boxes} are ground truths and the \textcolor{green}{green boxes} are predictions from SamLP.}
	\label{fig:ccpd}
	\vspace{-8pt}
\end{figure}

The results on UFPR-ALPR \cite{UFPR-ALPR} are illustrated in Fig. \ref{fig:ufpr} and Table \ref{tab:LP_UFPR}. From Table \ref{tab:LP_UFPR}, we find that the proposed SamLP and SamLP\_P all achieve better detection accuracy compared to other recent LP detectors. When we only apply LoRA fine-tuning in SAM \cite{SAM}, the proposed SamLP reaches 96.8 on F1-score and 94.9 on AP which has already exceeds recent popular LP detectors. And the promptable fine-tuning in SamLP\_P maintains the promptable ability of SAM \cite{SAM} and further improve the detection accuracy, increasing by 0.1 on F1-score and 0.3 on AP. This improvement may benefit by the extra training in promptable fine-tuning which additionally promotes the performance of mask decoder in SamLP.

The detection performance on CCPD \cite{CCPD} is illustrated in Fig. \ref{fig:ccpd} and Table \ref{tab:LP_CCPD}. All LP detection methods in Table \ref{tab:LP_CCPD} are trained on CCPD-Base subset and tested on all subsets in CCPD \cite{CCPD} to get the AP metric. On the testing samples from CCPD-Base, our proposed SamLP and SamLP\_P all achieve the best detection performance (99.9 AP). For other degenerated subsets, our proposed SamLP and SamLP\_P reach the most of best performance. Only the performance on CCPD-DB subset is weak compared to other LP detectors. This may because our foundation model, \emph{i.e.} SAM \cite{SAM}, is weak on segmentation under low-light scenes due to lack of low-light training data in SA-1B \cite{SAM}. Although the SAM \cite{SAM} is unsuitable for low-light scenes, our proposed SamLP also obtain 91.8 AP on CCPD-DB which is the second place in compared detectors.

\begin{figure}[h]
	\centering
	\subfloat[CRPD\_single]{
		\label{fig:crpd_s}
		\includegraphics[width=0.49\linewidth]{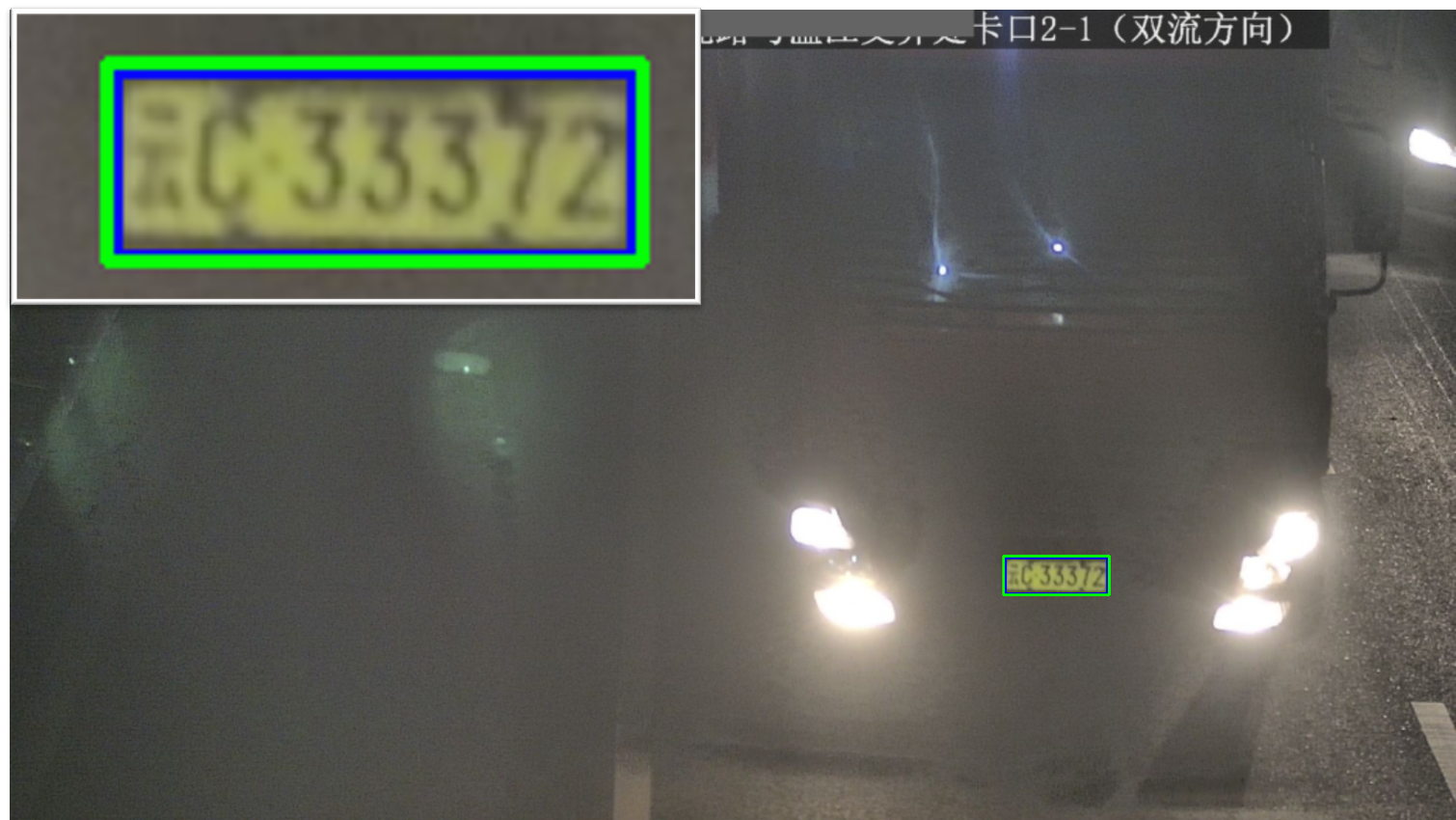}
		\includegraphics[width=0.49\linewidth]{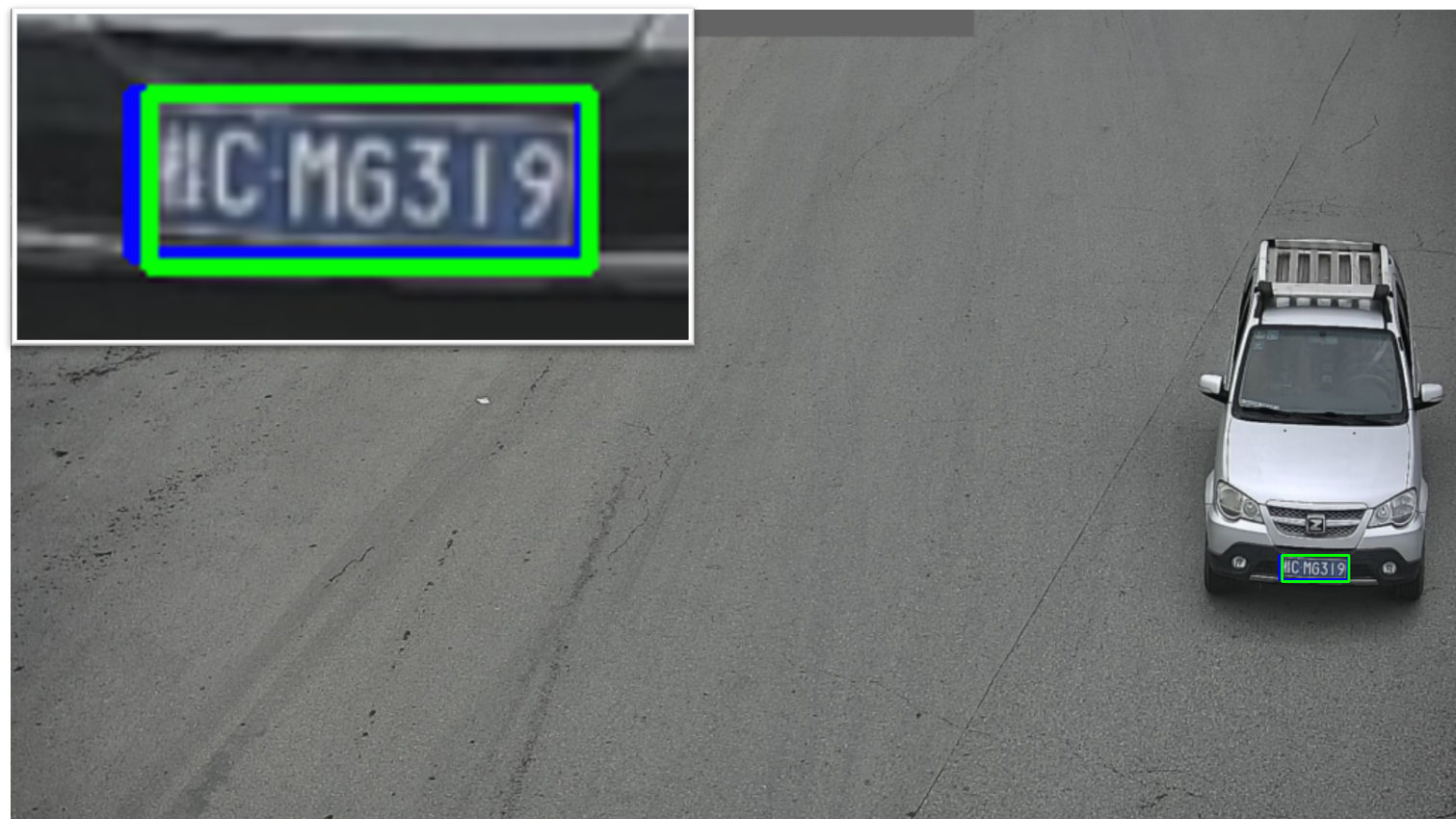}
	}
	\hfil
	\subfloat[CRPD\_multi]{
		\label{fig:crpd_m}
		\includegraphics[width=0.49\linewidth]{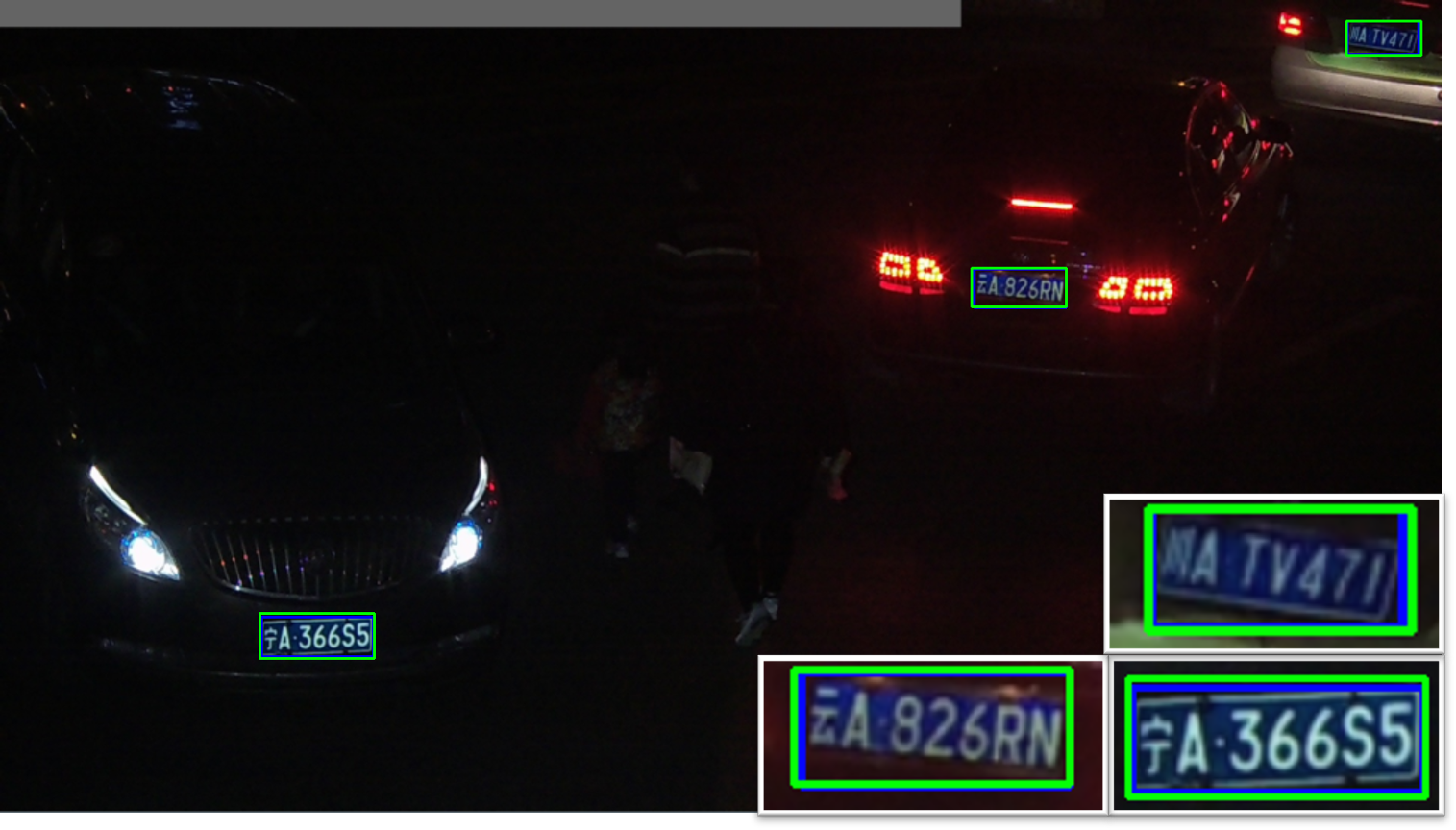}
		\includegraphics[width=0.49\linewidth]{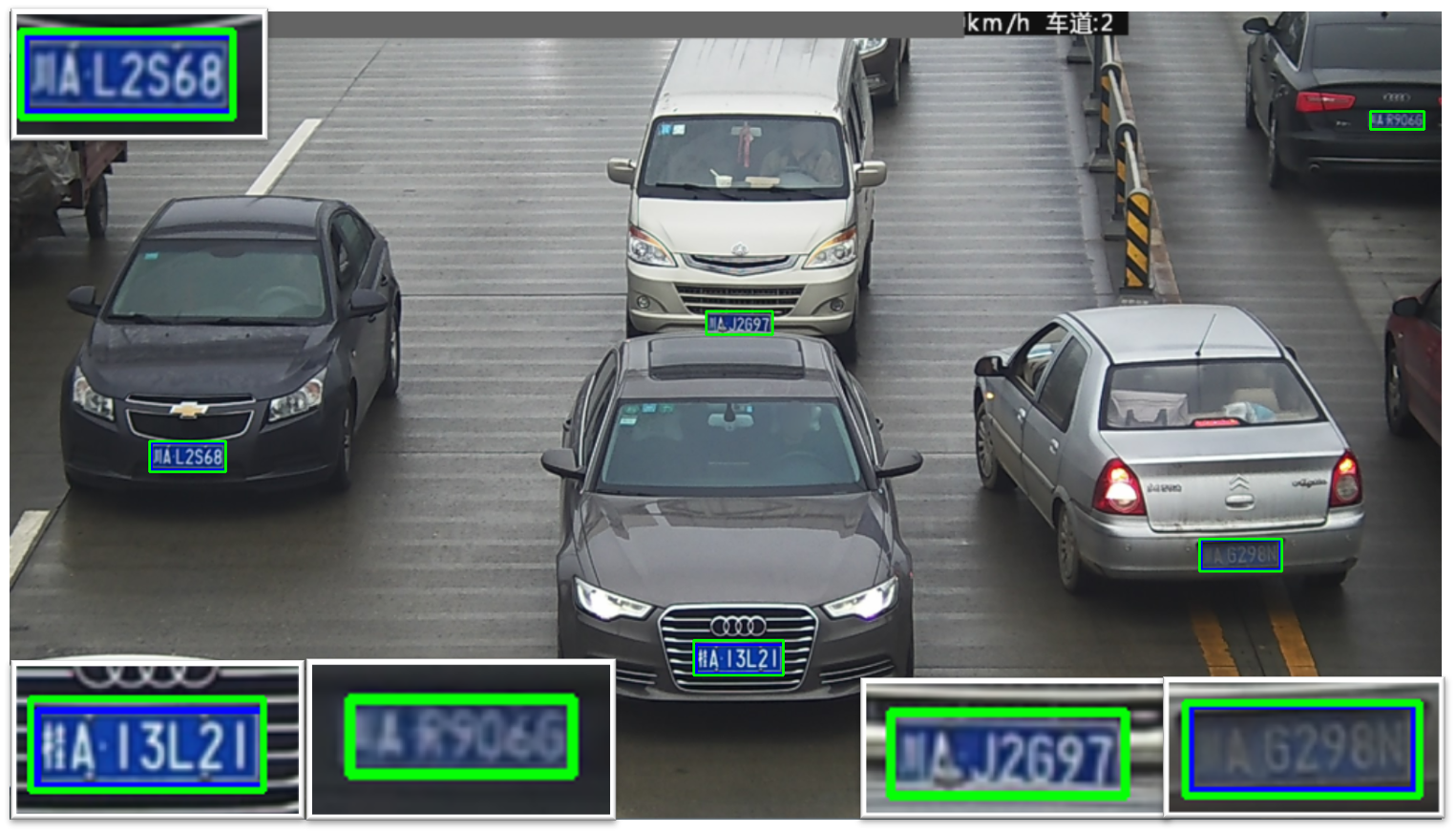}
	}
	\caption{Some detection results of our proposed SamLP on CRPD \cite{CRPD}. The \textcolor{blue}{blue boxes} are ground truths and the \textcolor{green}{green boxes} are predictions from SamLP.}
	\label{fig:crpd}
	\vspace{-10pt}
\end{figure}

The detection performance on CRPD \cite{CRPD} is shown in Fig. \ref{fig:crpd} and Table \ref{tab:LP_CRPD}. We compare the detection performance on two typical subsets from CRPD \cite{CRPD}, \emph{i.e.} CRPD\_single and CRPD\_multi. Every single image in CRPD\_single only contains a single LP instance, and a single image in CRPD\_multi contains more than two LP instances which is more challenging in detection. As for single LP detection in CRPD\_single, our proposed SamLP achieves competitive performance compared to recent LP detectors or scene text detectors (98.0 on F1-score). And the promptable fine-tuning (\emph{i.e.} SamLP\_P) obtain significant improvement, reaching the best detection accuracy (99.0 on F1-score). As for the detection of multiple LPs in CRPD\_multi, the SamLP and SamLP\_P all outperform other compared LP detectors. The CRPD\_multi subset includes more LPs and its street scene is more complex which means it is more challenging than CRPD\_single. The better performance of proposed SamLP shows it has great potential for extracting discriminative LP features and getting accurate LP detection performance. 

\subsubsection{Few-Shot Performance} The SAM \cite{SAM} is trained on a huge dataset and it has excellent generalization ability which can be transfer to diverse downstream tasks. Benefiting by the generalization ability, the foundation models are considered as outstanding few-shot learner \cite{GPT3}. Therefore, we attempt to explore the few-shot learning ability of SamLP. We only select a small proportion of training set to train the SamLP and test it on the whole testing set. The few-shot learning performance on UFPR-ALPR \cite{UFPR-ALPR} is shown in Table \ref{tab:few}.


\begin{table*}
	\caption{The few-shot learning ability of proposed SamLP. All models are trained with partial training data and directly tested on whole testing data.}
	\centering
	\begin{tabular}{m{0.15\linewidth}|c|c|cc|cc|cc|cc}
		\toprule[1pt]
		\multirow{2}*{Method}  & \multirow{2}*{Dataset} &  \multirow{2}*{Promptable} & \multicolumn{2}{c|}{ 3\% (60 images)} & \multicolumn{2}{c|}{ 10\% (180 images)} & \multicolumn{2}{c|}{ 30\% (540 images)} & \multicolumn{2}{c}{ 100\% (1800 images)}  \\
		\cmidrule[1pt]{4-11}
		& & &  F1-score & AP &  F1-score & AP &  F1-score & AP &  F1-score & AP  \\
		\midrule[1pt]
		YOLOv3 \cite{YOLOv3} & \multirow{9}*{UFPR-ALPR} & \multirow{9}*{\ding{56}} &- & 72.8 & -  & 81.3 &  -& 86.2  & - & 87.8 \\
		YOLOX \cite{YOLOX}& & &- & 76.0 & -  & 88.4 &  -& 93.2 & - & 94.3 \\
		YOLOF \cite{YOLOF} & & &- & 77.9 & -  & 89.0 &  -& 93.7 & - & 93.7 \\
		RetinaNet \cite{RetinaNet} & & &- & 71.4 & -  & 78.3 &  -& 85.6  & - & 93.9 \\
		FCOS \cite{FCOS} & & &- & 55.8 & -  & 60.2 &  -& 79.0 & - & 92.9 \\
		DETR \cite{DETR} & & &- & 0.0 & -  & 0.0 &  -& 0.0 & - & 0.0 \\
		Deformable DETR \cite{DeformDETR} & & &- & 44.4 & -  & 51.4  &  -& 65.6 & - & 73.7 \\
		Conditional DETR \cite{CondDETR} & & &- & 0.0 & -  & 0.0 &  -&  0.3 & - & 0.4 \\
		DAB-DETR \cite{DABDETR} & & &- & 0.0 & -  & 0.0 &  -& 0.0 & - & 77.0 \\
		\midrule[1pt]
		SamLP  & \multirow{2}*{UFPR-ALPR} & \ding{56} &  \textbf{94.8} & \textbf{91.3} &  95.3 & 92.3 & 96.7 & 94.8 &  96.8 &94.9 \\
		SamLP\_P & & \ding{52}  &  94.1  & 90.2 & \textbf{96.1} & \textbf{92.7}  & \textbf{96.8} & \textbf{95.0} &  \textbf{96.9} & \textbf{95.2}  \\
		\bottomrule[1pt]
	\end{tabular}

	\vspace{2pt}
	\footnotesize{The compared detection methods are reimplemented through open source object detection toolbox MMdetection \mbox{\cite{mmdetection}}.\\}
	\label{tab:few}
\end{table*}

From Table \ref{tab:few}, we find that using extremely few training data (only 3\% or 60 images from UFPR-ALPR \cite{UFPR-ALPR}) for the training of SamLP has already achieved great LP detection performance. Under this condition, our proposed SamLP and SamLP\_P still achieve 91.3 AP and 90.2 AP respectively. But the detection performance of compared LP detectors decrease dramatically due to the lack of training data. 

The LP detection accuracy of all methods raise when the amount of training data increases. This means that the amount of training data is important for traditional deep learning methods. As for foundation model, the influence from data is lesser than traditional neural networks due to the pre-training on big data. Thus, the proposed SamLP and SamLP\_P maintain high detection accuracy (higher than 90 F1-score and 90 AP) under all proportion of training data and achieve best detection performance on $100\%$ of training data (96.9 F1-score and 95.2 AP). On the contrary, only after the training data are more than $30\%$ of training set, some traditional LP detectors reach detection accuracy higher than 90 AP, \emph{i.e.} 93.2 AP for YOLOX \cite{YOLOX} and 93.7 AP for YOLOF \cite{YOLOF} under $30\%$ of training data. Under $100\%$ of training data, the traditional LP detectors still can not exceed the proposed SamLP. Meanwhile, the performance of promptable fine-tuning also benefits from the increasing of training data amount. We find that SamLP\_P outperforms SamLP when the proportion of training data is more than $10\%$. This may because extremely few training data can not provide diverse prompt points for fine-tuning and cause the insufficient training of promptable segmentation. 

In addition, the training of transformer extremely relies on data. Wang \emph{et al.} \cite{DBLP:conf/eccv/WangZCST22} find that the detection transformer has serious dependency on rich training data. This is the reason why DETR \cite{DETR}, Deformable DETR \cite{DeformDETR}, Conditional DETR \cite{CondDETR}, and DAB-DETR \cite{DABDETR} reach extremely low accuracy (even 0.0 AP) on few-shot learning experiments. The proposed SamLP is also a transformer-based method, but it can overcome this drawback of transformer and obtain high accuracy. This means fine-tuning a foundation model into downstream tasks alleviates the data dependency and makes the training of transformer on limited data possible. Because of the adequate pre-training of SAM \cite{SAM}, the proposed SamLP gains outstanding few-shot learning ability. This shows the generous potential of foundation model on LP detection task.

\begin{figure}[h]
	\centering
	\subfloat[Training on UFPR \cite{UFPR-ALPR} and testing on CRPD\_single \cite{CRPD}]{
		\label{fig:ufpr2crpd}
		\includegraphics[width=0.49\linewidth]{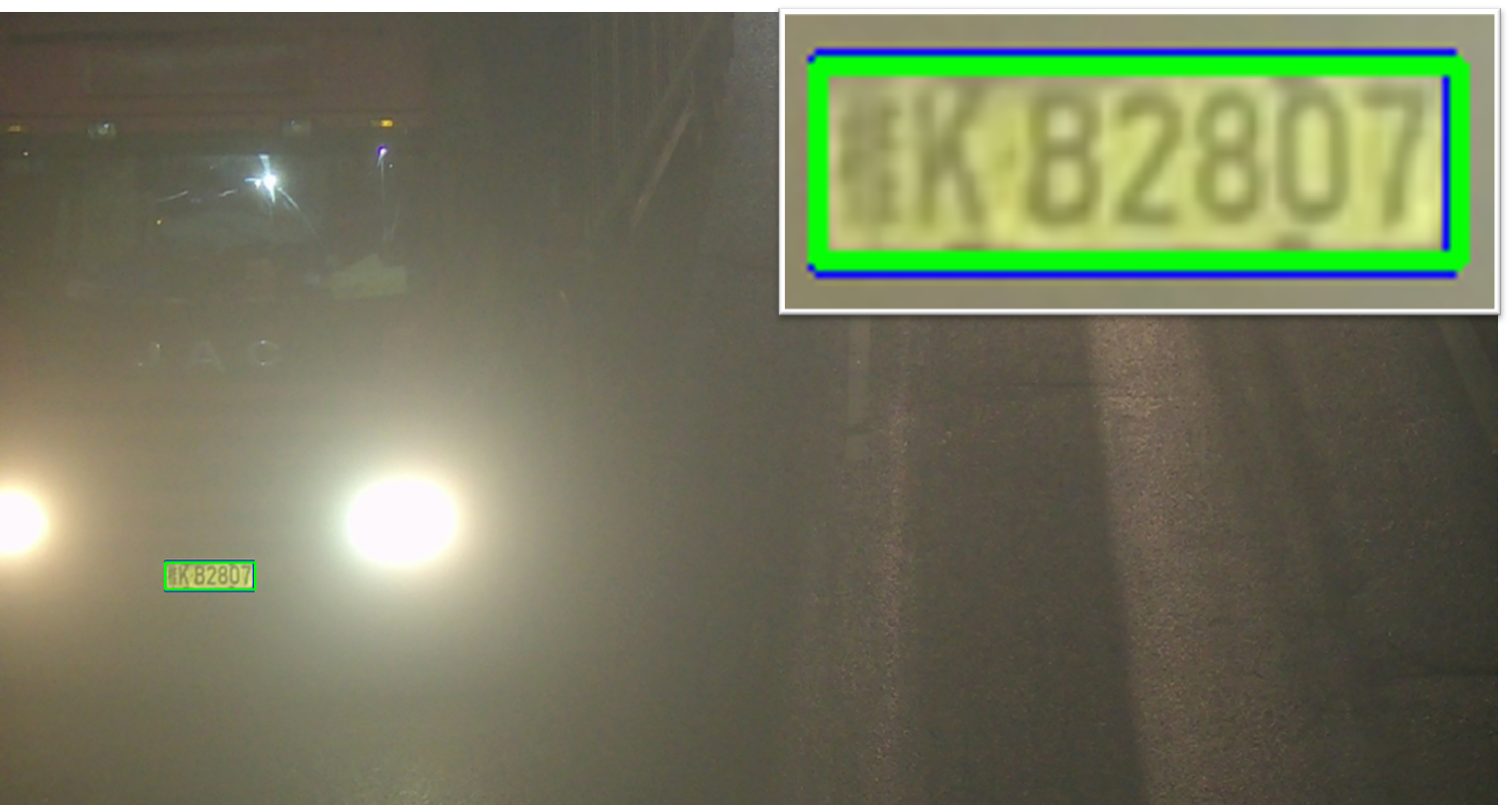}
		\includegraphics[width=0.49\linewidth]{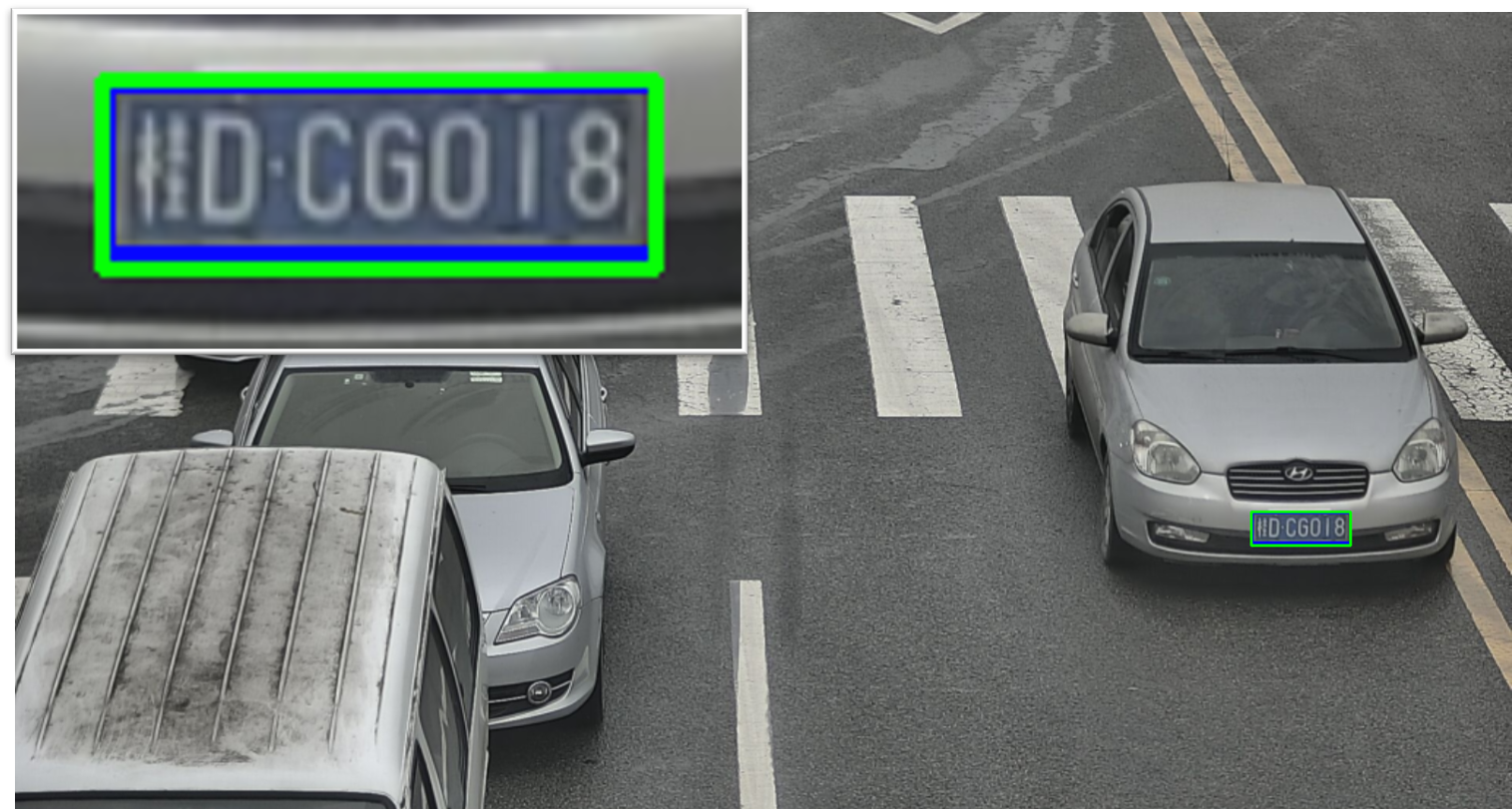}
	}
	\hfil
	\subfloat[Training on CCPD \cite{CCPD} and testing on AOLP \cite{AOLP}]{
		\label{fig:ccpd2aolp}
		\includegraphics[width=0.49\linewidth]{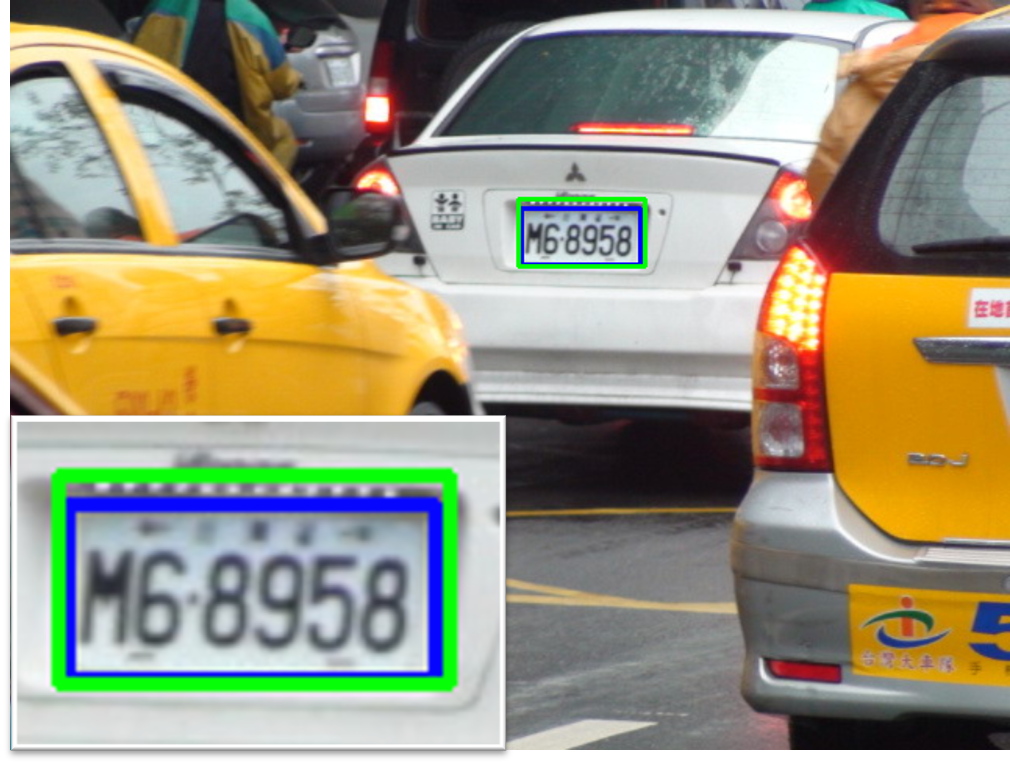}
		\includegraphics[width=0.49\linewidth]{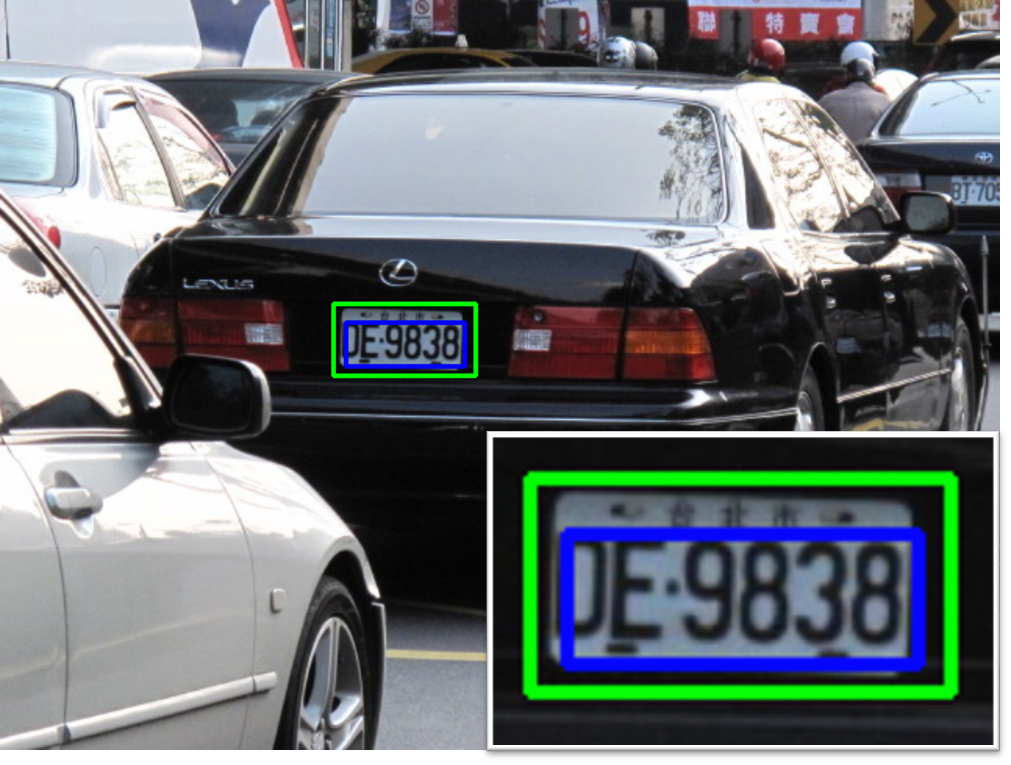}
	}
	\caption{Some zero-shot detection results of our proposed SamLP. The \textcolor{blue}{blue boxes} are ground truths and the \textcolor{green}{green boxes} are predictions from SamLP.}
	\label{fig:zero}
\end{figure}
\subsubsection{Zero-Shot Performance} As a unique identity of vehicles, LP is a special sign in traffic. Different countries and regions have different LPs. The LPs have diverse styles and contents, which is a significant characteristic in LP detection. As for traditional LP detectors, a special detectors is commonly trained by a special LP dataset, \emph{e.g.} UFPR-ALPR \cite{UFPR-ALPR} for Brazilian LPs, CCPD \cite{CCPD} for Chinese LPs, \emph{etc.} This leads the traditional LP detectors only suit for the detection on LPs which have the same style as training data. For example, the LP detector trained on Brazilian LP datasets may not suit for Chinese LP detection task because of the differences on appearance and style. Recent researches have already shown that the foundation models have excellent zero-shot transfer ability \cite{GPT3,CLIP}. Thus, with the help of transferable vision foundation model SAM \cite{SAM}, the proposed SamLP has better zero-shot transfer ability than traditional deep learning methods. To explore the zero-shot performance of SamLP, we train the LP detectors on UFPR-ALPR \cite{UFPR-ALPR} dataset (Brazilian LPs) but test them on CRPD \cite{CRPD} dataset (Chinese LPs) and train detectors on CCPD \cite{CCPD} but test them on AOLP \cite{AOLP}. The zero-shot detection performance is shown in Table \ref{tab:zero}.

\begin{table}[h]
	\caption{The zero-shot detection performance of proposed SamLP. All models are trained on LPs with a special style and directly tested on LPs with other styles.}
	\centering
	\begin{tabular}{m{0.31\linewidth}|m{0.1\linewidth}<{\centering}|m{0.1\linewidth}<{\centering}|c|m{0.02\linewidth}<{\centering} m{0.03\linewidth}<{\centering}}
		\toprule[1pt]
		Method & Training & Testing & Promptable & F1 & AP \\
		\midrule[1pt]
		YOLOv3 \cite{YOLOv3} & \multirow{8}*{\thead{UFPR-\\ALPR}} & \multirow{8}*{\thead{CRPD\_\\single}} &\multirow{8}*{\ding{56}}  & - & 77.4 \\
		YOLOX \cite{YOLOX} & & & & - & 85.7 \\
		YOLOF \cite{YOLOF} & & & & - & 81.7 \\
		RetinaNet \cite{RetinaNet} & & & & - & 78.4 \\
		FCOS \cite{FCOS} & & & & - & 41.3 \\
		Deformable DETR \cite{DeformDETR} & & & & - & 49.9 \\
		CLPD \cite{CLPD} & & & & - & 61.9 \\
		SCCA \cite{SCCA} & & & & - & 77.7  \\
		\midrule[1pt]
		 SamLP  & \multirow{2}*{\thead{UFPR-\\ALPR}} & \multirow{2}*{\thead{CRPD\_\\single}} & \ding{56} & \textbf{94.3} & \textbf{90.3}  \\
		 SamLP\_P  &  &  & \ding{52} & 93.8 & \textbf{90.3}  \\
		\midrule[1pt]
		YOLOX \cite{YOLOX} & \multirow{6}*{\thead{CCPD}} &  \multirow{6}*{\thead{AOLP\_\\LE}} & \multirow{6}*{\ding{56}} & - & 9.3  \\
		YOLOF \cite{YOLOF} & & & & - & 1.0  \\
		FCOS \cite{FCOS} & & & & - & 1.0  \\
		Deformable DETR \cite{DeformDETR} & & & & - & 0.0   \\
		CLPD \cite{CLPD} & & & & - & 19.6 \\
		SCCA \cite{SCCA} & & & & - & 0.2  \\
		\midrule[1pt]
		SamLP  & \multirow{2}*{\thead{CCPD}} & \multirow{2}*{\thead{AOLP\_\\LE}} & \ding{56} &  \textbf{90.3} & \textbf{84.3}  \\
		SamLP\_P  &  &  & \ding{52} & 82.7  & 70.5    \\
		\bottomrule[1pt]
	\end{tabular}

	\vspace{2pt}
	\footnotesize{The compared detection methods are reimplemented through open source object detection toolbox MMdetection \mbox{\cite{mmdetection}}.\\}
	\label{tab:zero}
\end{table}


We find that the traditional LP detectors have many false predictions in the testing on CRPD \cite{CRPD} dataset. As for single LP detection, the best performance of traditional LP detectors is only 85.7 AP reached by YOLOX \cite{YOLOX}, and our proposed SamLP and SamLP\_P achieve 90.3 AP easily which has significant improvement compared to traditional LP detectors. Meanwhile, we select the most challenging subset in AOLP \cite{AOLP} (LE subset denoted as AOLP\_LE) to test the LP detectors trained on CCPD \cite{CCPD}. Our proposed SamLP achieves best zero-shot detection performance (84.3 AP) on AOLP\_LE \cite{AOLP}. The compared LP detectors only reach the best accuracy of 19.6 AP (CLPD \cite{CLPD}) which means the zero-shot transfer ability of compared LP detectors are limited. These results demonstrate the excellent zero-shot transferring ability of SamLP. The reason why SamLP\_P do not outperform SamLP is because the promptable fine-tuning make SamLP concentrate on the LP features from UFPR-ALPR \cite{UFPR-ALPR} or CCPD \cite{CCPD} dataset deeper. This is harmful to the detection on CRPD \cite{CRPD} and AOLP \cite{AOLP} datasets because there is domain difference between train set and test set. 

In summary, the proposed SamLP and SamLP\_P all achieve better detection accuracy on LP datasets from different countries and regions, which means the proposed methods have stable performance on diverse LPs. In comparison between SamLP and SamLP\_P, we find that the promptable fine-tuning (\emph{i.e.} SamLP\_P) relies more on diverse data than LoRA fine-tuning (\emph{i.e.} SamLP), which means the proposed SamLP are better few-shot learner and more suitable for circumstance under data scarcity compared to SamLP\_P. And the zero-shot detection performance also claims that the SamLP has better zero-shot performance compared to SamLP\_P. This is because SamLP\_P focuses on the detection of specific LPs due to the deep promptable fine-tuning on specific dataset, limiting the generalization ability of SamLP\_P. Although the SamLP\_P has better detection accuracy on some experiments, the proposed SamLP has better stability and robustness on LP detection.

\subsection{Ablation Study}\label{sec:exp:abl}

To further analyze the fine-tuning of SAM \cite{SAM} on LP detection, some ablation studies are implemented in this section to explore the better configuration of proposed SamLP. 

\subsubsection{LoRA Layers}
\begin{table}[h]
	\caption{The influence of where the LoRA layers injected.}
	\centering
	\begin{tabular}{m{0.12\linewidth}<{\centering}|m{0.12\linewidth}<{\centering}|m{0.15\linewidth}<{\centering}|m{0.06\linewidth}<{\centering} m{0.06\linewidth}<{\centering} m{0.06\linewidth}<{\centering} m{0.06\linewidth}<{\centering}}
		\toprule[1pt]
		Image Encoder & Mask Decoder & \# of Trainable Param  & P & R & F1 & AP \\
		\midrule[1pt]
		\ding{52} & & 0.14 M & 94.7 & 98.7 & 96.6 & 94.2 \\
		 & \ding{52} & 0.02 M & 93.2 & 94.9 & 94.1 & 89.3 \\
		\ding{52} & \ding{52} & 0.17 M & 95.3 & 98.4 & 96.8 & 94.9 \\
		\bottomrule[1pt]
	\end{tabular}
	\label{tab:lora_layers}
\end{table}
We first explore where the LoRA layers should be injected in SAM \cite{SAM}. SAM \cite{SAM} mainly contains two transformer-based architectures, \emph{i.e.} image encoder and mask decoder. Thus, the LoRA layers can be injected in image encoder and mask decoder. Table \ref{tab:lora_layers} shows the performance about different injections. When we inject LoRA layers in image encoder, SamLP reaches 94.2 AP on UFPR-ALPR \cite{UFPR-ALPR}. But the injection in mask decoder only obtains 89.3 AP. This obviously shows that the image encoder plays the critical role in SAM \cite{SAM}. When we inject LoRA layers in both image encoder and mask decoder, SamLP achieves best performance (94.9 AP). Therefore, we decide inject LoRA layers in both image encoder and mask decoder in proposed SamLP.

\subsubsection{Prompt Types}

\begin{table}[h]
	\caption{The influence of prompt types used in SamLP and SamLP\_P.}
	\centering
	\begin{tabular}{m{0.12\linewidth}|m{0.12\linewidth}<{\centering}|c|m{0.08\linewidth}<{\centering} m{0.08\linewidth}<{\centering} m{0.08\linewidth}<{\centering} m{0.08\linewidth}<{\centering}}
		\toprule[1pt]
		Method &Point & Mask  & P & R & F1 & AP \\
		\midrule[1pt]
		\multirow{4}*{SamLP}&  & & 95.3 & 98.4 & \textbf{96.8} & \textbf{94.9} \\
		& \ding{52} & & 84.5 & 94.6 & 89.3 & 80.4 \\
		&& \ding{52} & 93.9 & 98.7 & 96.2 & 93.0 \\
		&\ding{52} & \ding{52} & 88.3 & 97.3  & 92.6  & 86.1   \\
		\midrule[1pt]
		\multirow{4}*{SamLP\_P}&  & & 94.1 & 97.6 & 95.8 & 92.7 \\
		&\ding{52} & & 94.3 & 98.3 & 96.3 & 94.3 \\
		&& \ding{52} & 95.3 & 98.6 & \textbf{96.9} & \textbf{95.2} \\
		&\ding{52} & \ding{52} & 93.8 & 98.8 & 96.3 & 94.4  \\
		\bottomrule[1pt]
	\end{tabular}
	\label{tab:prompt_type}
\end{table}

In Section \ref{sec:method:prompt}, we claim that there are diverse prompt types in SAM \cite{SAM}, \emph{i.e.} point, box, mask. We select point prompt and mask prompt in training because the point prompt is the most flexible prompt in SAM \cite{SAM} and the mask prompt has riches image information. In inference, two types of prompt can be used in SamLP, \emph{i.e.} point and mask. 

In Table \ref{tab:prompt_type}, we analyze the influence of prompt types in SamLP during inference. We only fine-tune the image encoder and mask decoder in SamLP, so that the promptable segmentation ability of SamLP is weak. And the experimental results also show that the detection performance decrease significantly when the prompts are introduced into SamLP. Therefore, the proposed SamLP is unsuitable for promptable segmentation, and it reaches best detection accuracy (96.8 F1-score and 94.9 AP) when there is no prompt. 

To obtain the promptable segmentation ability in SamLP, we propose promptable fine-tuning in Section \ref{sec:method:prompt}. After promptable fine-tuning, SamLP\_P achieves better promptable segmentation performance. Using point as prompt, the SamLP\_P reaches 96.3 F1-score and 94.3 AP. Using mask as prompt, SamLP\_P achieves best detection performance (96.9 F1-score and 95.2 AP). And when we use both point and mask as prompts, the proposed SamLP\_P attains 96.3 F1-score and 94.4 AP. SamLP\_P achieves better performance when only mask prompt is used. This is because the mask prompt introduce richest image content information into SamLP. And the initial prompt points are sampled from prediction with $\mathrm{None}$ prompt which may has false prediction, leading to the errors in refinement with prompt. Therefore, introducing point prompt into SamLP\_P causes the reduction of detection accuracy. Finally, we select $\mathrm{None}$ prompt in SamLP and mask prompt in SamLP\_P in inference.

\subsubsection{LoRA Rank}
\begin{figure}[h]
	\includegraphics[width=1.0\linewidth]{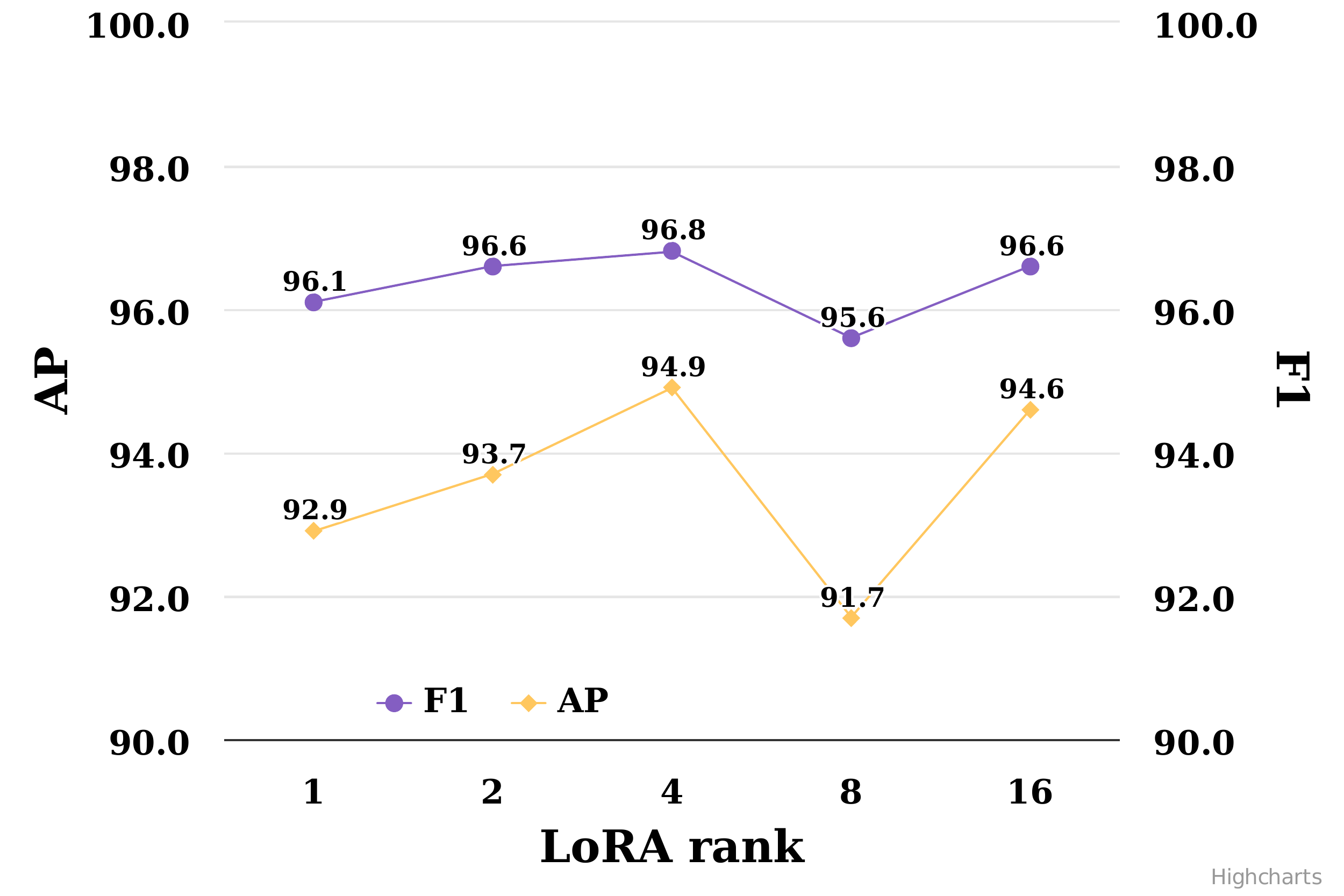}
	\centering
	\caption{The influence of LoRA rank $r$ in SamLP}
	\label{fig:LoRA_rank}
\end{figure}
The crucial parameter in LoRA layer is LoRA rank $r$. Different LoRA rank leads to different detection performance. Fig. \ref{fig:LoRA_rank} demonstrates the influence of LoRA rank $r$. We discover the detection performance of proposed SamLP raises continuously in a certain range of LoRA rank. It is obvious that too small rank size can not adapt SAM \cite{SAM} to LP detection. After $r=4$, the detection accuracy of SamLP starts fluctuation. A large rank size $r$ causes the increase of the number of training parameters but there is no significant improvement compared to $r=4$. Thus, to balance the model size and detection performance, we finally decide $r=4$ in proposed SamLP.

\begin{figure}[h]
	\includegraphics[width=1.0\linewidth]{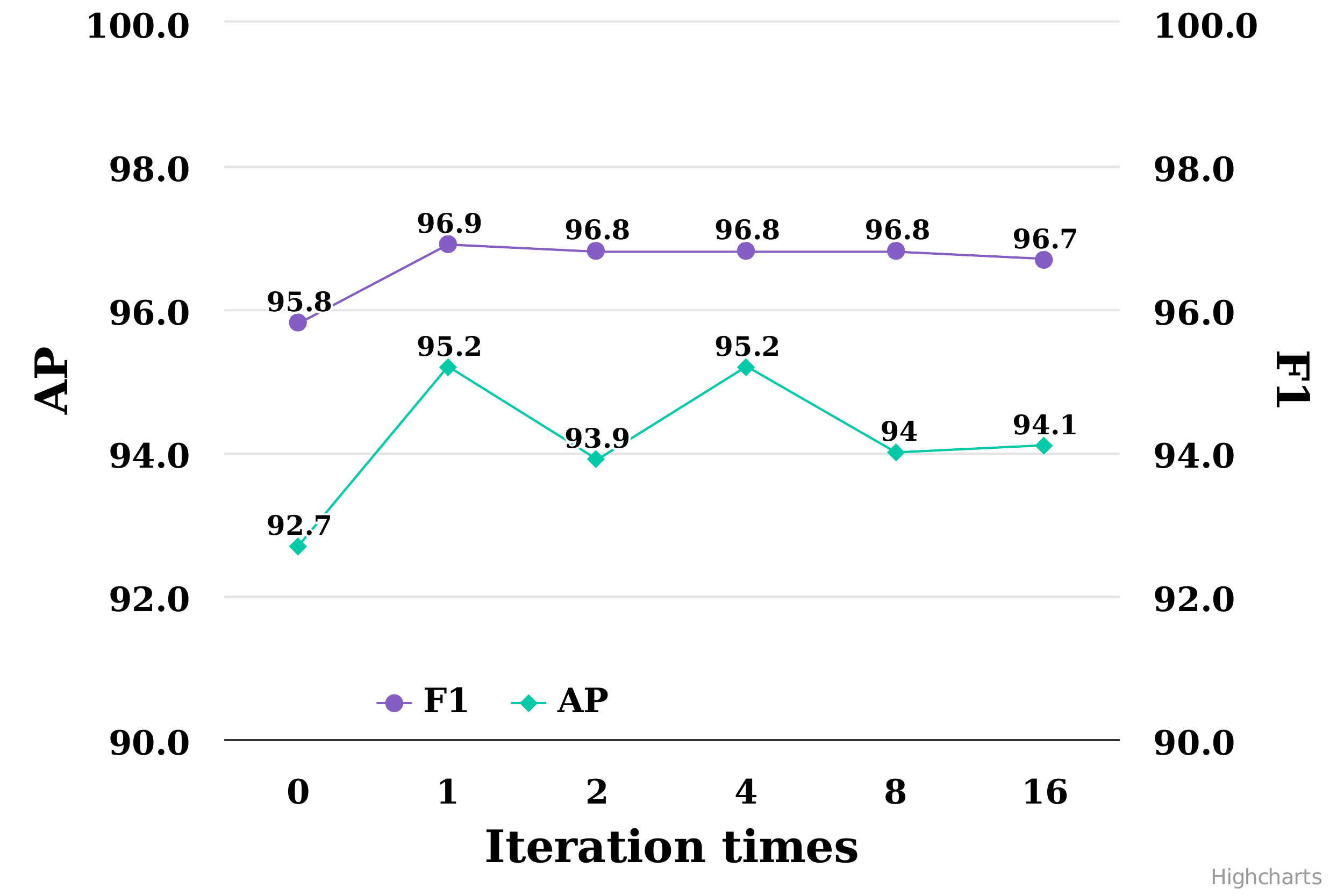}
	\centering
	\caption{The influence of iteration times $\mathrm{Num}$ in refinement of SamLP\_P}
	\label{fig:refine}
\end{figure}

\subsubsection{Refinement Iteration}
In Section \ref{sec:exp:imp}, we claim that SamLP\_P uses iterative refinement to further improve the detection performance in inference. We explore the refinement iteration times $\mathrm{Num}$ in Fig. \ref{fig:refine} to verify the cascade iterative refinement in promptable segmentation is effective. When there is no cascade refinement (\emph{i.e.} $\mathrm{Num}$=0), the detection accuracy of proposed SamLP\_P is lower than those with refinement. With the increase of refinement iteration times $\mathrm{Num}$, the detection accuracy is fluctuating and reaches the highest value at $\mathrm{Num}=1$ and $\mathrm{Num}=4$. We finally choose $\mathrm{Num}=1$ in SamLP\_P because fewer refinement iteration consumes less time in inference.



\section{Conclusion}\label{sec:con}

This paper proposes a customized Segment Anything Model (SAM) \cite{SAM} for license plate detection task, named SamLP. The proposed SamLP is the first LP detector based on vision foundation model and it achieves promising detection performance on several LP detection dataset. Specifically, we introduce a parameter-efficient fine-tuning method called LoRA into a powerful vision foundation model, \emph{i.e.} SAM \cite{SAM} to adapt SAM \cite{SAM} into LP detection task. In SamLP, we inject LoRA layers into image encoder and mask decoder of SAM \cite{SAM} and the LoRA fine-tuning makes SAM \cite{SAM} concentrate on LP detection successfully. Furthermore, to maintain the promptable fine-tuning capacity of SAM \cite{SAM}, we design a promptable fine-tuning step to further fine-tune the SamLP and realize the SamLP\_P. Comparing the performances of SamLP\_P and SamLP, the proposed promptable fine-tuning step effectively improve the promptable segmentation capacity. The experiments demonstrate that the proposed SamLP outperforms the popular LP detectors. Meanwhile, the proposed SamLP has better few-shot and zero-shot learning ability compared to other LP detectors. These all show that the potential and advantage of vision foundation models on LP detection task.

In this paper, we attempt transfer recent popular vision foundation model into LP detection task. However, recent vision foundation models all contain a huge number of parameters which brings heavy burden in computation and causes slow inference speed in application. In the future, we will attempt to introduce the knowledge distillation method into vision foundation model to reduce the inference time and improve the efficiency of LP detector based on vision foundation model.

 

\bibliography{ref}
\bibliographystyle{IEEEtran}

\begin{IEEEbiography}[{\includegraphics[width=1in,height=1.25in,clip,keepaspectratio]{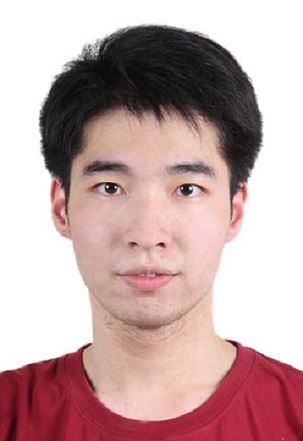}}]{Haoxuan Ding} received the B.E. degree and the M.S. degree in aerospace propulsion theory and engineering from the Northwestern Polytechnical University, Xi’an, China, in 2018 and 2021 respectively. He is currently pursuing the Ph.D. degree from Unmanned System Research Institute, Northwestern Polytechnical University, Xi’an, China. His research interests include computer vision and pattern recognition.
\end{IEEEbiography}

\begin{IEEEbiography}[{\includegraphics[width=1in,height=1.25in,clip,keepaspectratio]{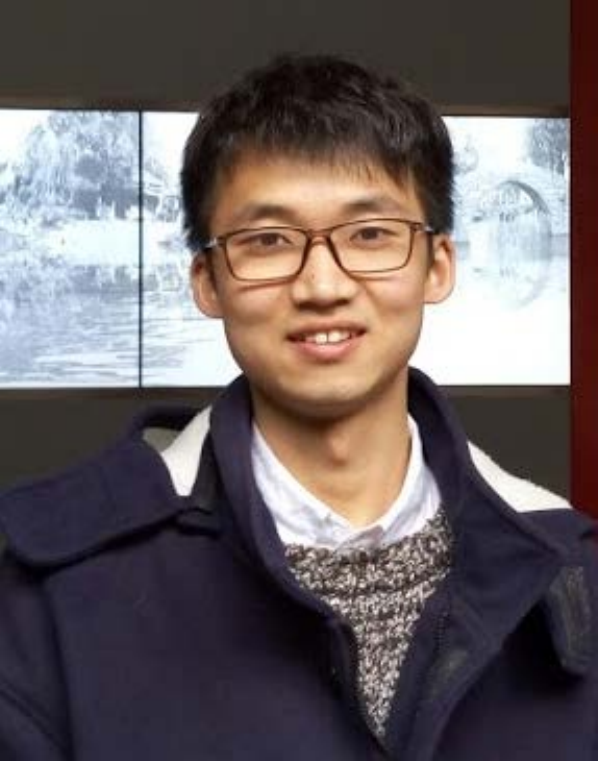}}]{Junyu Gao} received the B.E. degree and the Ph.D. degree in computer science and technology from the Northwestern Polytechnical University, Xi’an	710072, Shaanxi, P. R. China, in 2015 and 2021 respectively. He is currently an associate professor with the School of Artificial Intelligence, Optics and Electronics (iOPEN), Northwestern Polytechnical University, Xi’an, China. His research interests include computer vision and pattern recognition.
\end{IEEEbiography}

\begin{IEEEbiography}[{\includegraphics[width=1in,height=1.25in,clip,keepaspectratio]{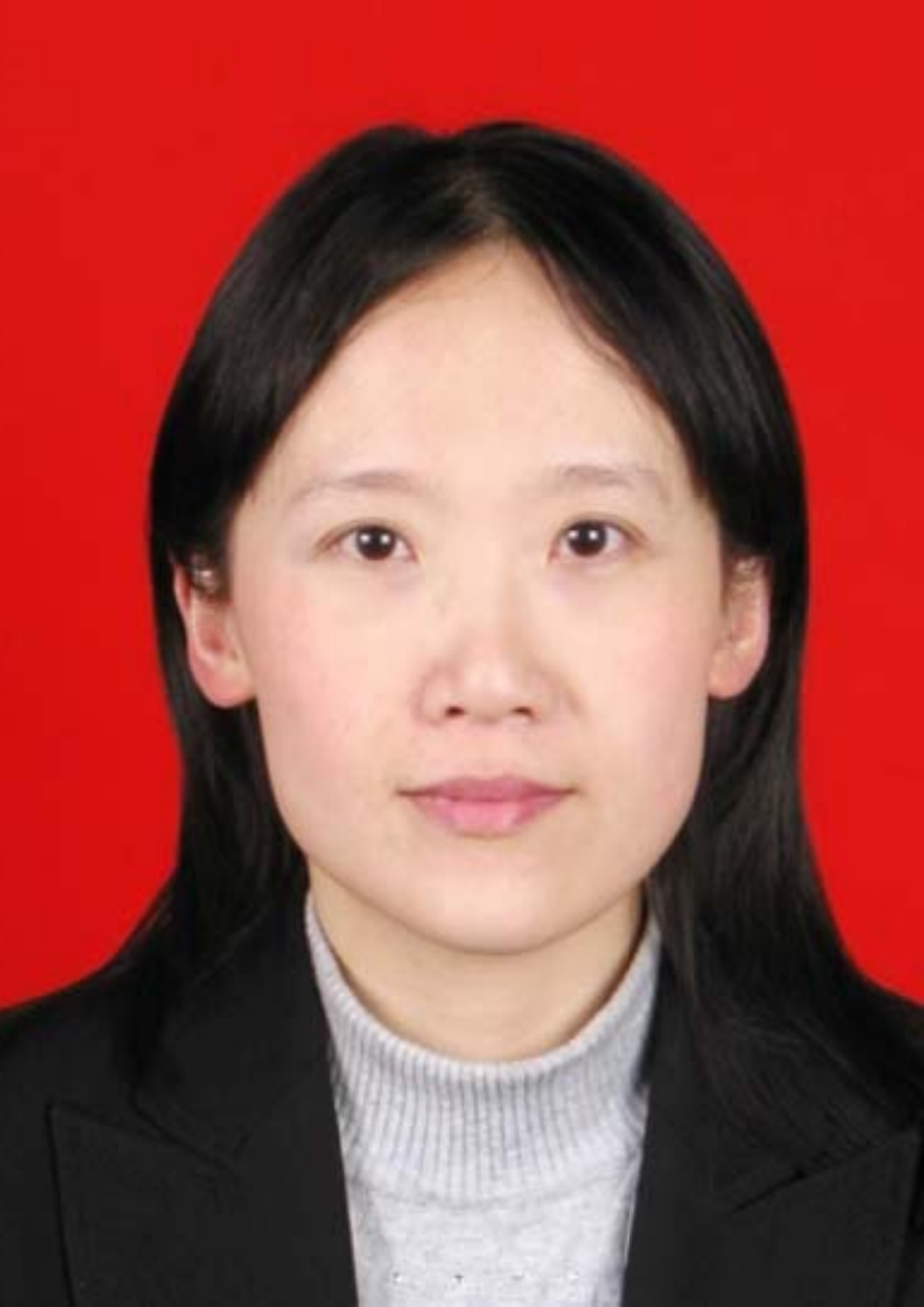}}]{Yuan Yuan}(M'05-SM'09) is currently a Full Professor with the School of Computer Science and the School of Artificial Intelligence, Optics and Electronics (iOPEN), Northwestern Polytechnical University, Xi’an, China. She has authored or co-authored over 150 papers, including about 100 in reputable journals, such as the IEEE TRANSACTIONS and PATTERN RECOGNITION, as well as the conference papers in CVPR, BMVC, ICIP, and ICASSP. Her current research interests include visual information processing and image/video content analysis.
\end{IEEEbiography}
%
\begin{IEEEbiography}[{\includegraphics[width=1in,height=1.25in,clip,keepaspectratio]{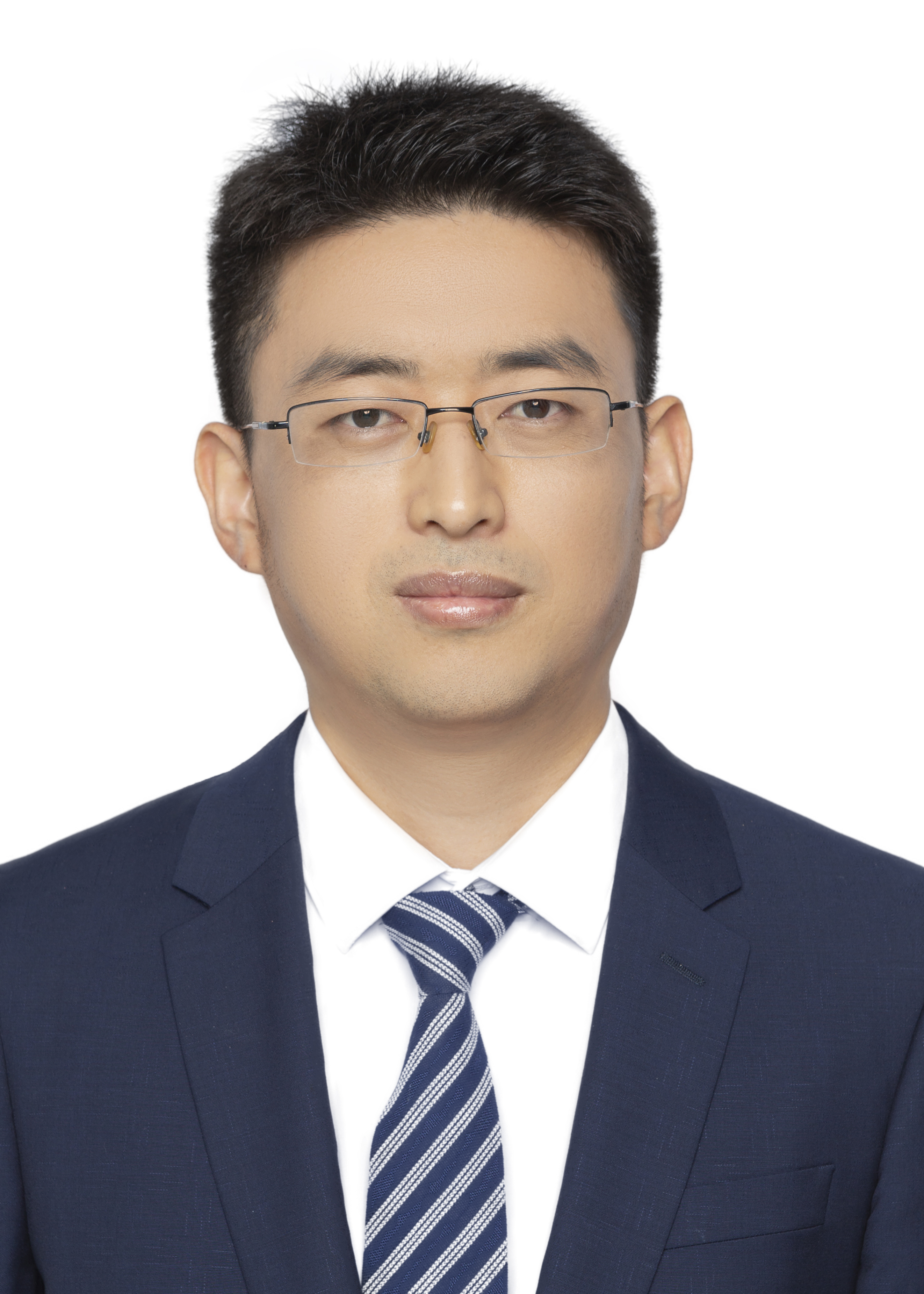}}]{Qi Wang}(M'15-SM'15) received the B.E. degree in automation and the Ph.D. degree in pattern recognition and intelligent systems from the University of Science and Technology of China, Hefei, China, in 2005 and 2010, respectively.  He is currently a Professor with the School of Artificial Intelligence, Optics and Electronics (iOPEN), Northwestern Polytechnical University, Xi'an, China. His research interests include computer vision, pattern recognition and remote sensing.
\end{IEEEbiography}
%
%
%

\vfill

\end{document}